\documentclass[runningheads]{llncs}
\usepackage{graphicx}

\usepackage{tikz}
\usepackage{comment}
\usepackage{amsmath,amssymb} 
\usepackage{color}
\usepackage{graphicx}
\usepackage{multirow}
\usepackage{booktabs}
\usepackage[abs]{overpic}
\usepackage{rotating}
\usepackage{sidecap}
\usepackage{caption}
\usepackage{hyperref}
\makeatletter
\newcommand\notsotiny{\@setfontsize\notsotiny{5.61415}{6.4828}}
\usepackage[accsupp]{axessibility}  
\usepackage{floatrow}
\usepackage{xcolor}

\begin{document}
\pagestyle{headings}
\mainmatter
\def\ECCVSubNumber{3644}  

\title{JPerceiver: Joint Perception Network for Depth, Pose and Layout Estimation in Driving Scenes} 

\titlerunning{JPerceiver: Joint Perception Network}
%
\author{Haimei Zhao\inst{1}\and
Jing Zhang\inst{1} \and
Sen Zhang\inst{1}
\and
Dacheng Tao\inst{2,1}}
\authorrunning{Haimei Zhao et al.}
%
\institute{The University of Sydney, 6 Cleveland St, Darlington, NSW 2008, Australia \and
JD Explore Academy, Beijing, China\\
\email{\{hzha7798,szha2609\}@uni.sydney.edu.au}\\
\email{jing.zhang1@sydney.edu.au
} \email{dacheng.tao@gmail.com}}

\maketitle

\begin{abstract}
Depth estimation, visual odometry (VO), and bird's-eye-view (BEV) scene layout estimation present three critical tasks for driving scene perception, which is fundamental for motion planning and navigation in autonomous driving. Though they are complementary to each other, prior works usually focus on each individual task and rarely deal with all three tasks together. A naive way is to accomplish them independently in a sequential or parallel manner, but there are three drawbacks, i.e., 1) the depth and VO results suffer from the inherent scale ambiguity issue; 2) the BEV layout is usually estimated separately for roads and vehicles, while the explicit overlay-underlay relations between them are ignored; and 3) the BEV layout is directly predicted from the front-view image without using any depth-related information, although the depth map contains useful geometry clues for inferring scene layouts. In this paper, we address these issues by proposing a novel joint perception framework named JPerceiver, which can simultaneously estimate scale-aware depth and VO as well as BEV layout from a monocular video sequence. It exploits the cross-view geometric transformation (CGT) to propagate the absolute scale from the road layout to depth and VO based on a carefully-designed scale loss. Meanwhile, a cross-view and cross-modal transfer (CCT) module is devised to leverage the depth clues for reasoning road and vehicle layout through an attention mechanism. JPerceiver can be trained in an end-to-end multi-task learning way, where the CGT scale loss and CCT module promote inter-task knowledge transfer to benefit feature learning of each task. Experiments on Argoverse, Nuscenes and KITTI show the superiority of JPerceiver over existing methods on all the above three tasks in terms of accuracy, model size, and inference speed. The code and models are available at~\href{https://github.com/sunnyHelen/JPerceiver}{https://github.com/sunnyHelen/JPerceiver}.
\end{abstract}

\begin{figure*}[!t]
 \begin{center}
  \includegraphics[height=2.8cm,width=10.6cm]{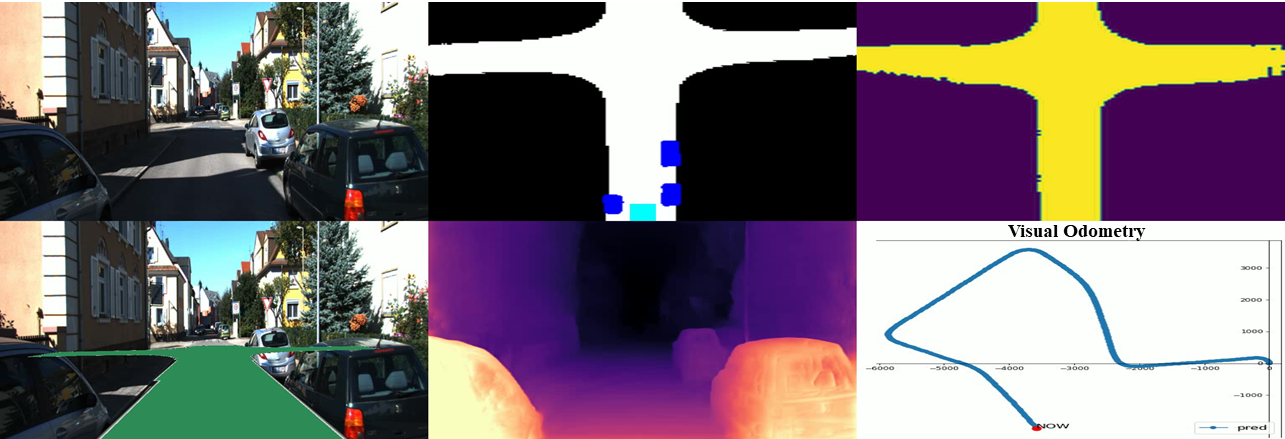} 
 \end{center}
 \caption{The proposed JPerceiver can predict BEV semantic layout (middle top, white for roads, blue for cars and cyan for the ego car), scale-aware depth map (middle bottom) and VO result (bottom right) simultaneously from a monocular video sequence (top left). The drivable area is visualized in green (bottom left) by projecting the BEV road layout onto the image plane. The ground truth BEV layout is shown in top right.}
\label{fig:openimage}
\end{figure*}

\section{Introduction}
Autonomous driving has witnessed great progress in recent years, where deep learning is playing an increasing role in perception \cite{Phillips_2021_CVPR,luo2021self,Zhuang_2021_ICCV}, planning \cite{Casas_2021_CVPR,Wang_2021_CVPR}, navigation \cite{Badki_2021_CVPR,Thavamani_2021_ICCV}, and decision making \cite{hang2020human,huang2021toward}. Among them, scene perception is the basis for other subsequent procedures in autonomous driving~\cite{zhang2020empowering}, which includes various sub-tasks for different perception purposes, e.g., depth estimation for 3D measurement \cite{fu2018deep,monodepth2}, ego motion estimation for localization and visual odometry (VO) \cite{wang2017deepvo,li2018undeepvo,zou2020learning,zhang2022towards}, as well as bird’s-eye-view (BEV) or front-view (FV) layout estimation for detecting obstacles and drivable areas \cite{mani2020monolayout,philion2020lift,yang2021projecting,chen2019progressive}. Although these tasks have underlying relations to each other intuitively, they are usually tackled separately in prior works \cite{monodepth2,yang2021projecting}. The joint estimation for all these tasks has not drawn enough attention so far, and the benefits and challenges of doing so remain unclear, which is the focus of this paper.

Depth and VO estimation are two closely related computer vision tasks that have been studied for decades \cite{torralba2002depth,nister2004visual}. Recent self-supervised learning methods use the photometric consistency between consecutive frames to achieve the simultaneous estimation of scene depth and VO from monocular video sequences, where no ground truth depth labels are required \cite{zhou2017unsupervised,monodepth2,shu2020featdepth}. On the other hand, BEV scene layout estimation refers to the task of estimating the semantic occupancy of roads and vehicles in the metric-scale BEV plane directly from FV images \cite{mani2020monolayout,philion2020lift,yang2021projecting,fiery2021}. Though significant progress has been made in each individual task, they still suffer from some inherent problems: i.e., (1) the scale ambiguity in monocular depth and VO estimation since the photometric error between corresponding pixels is equivalent up to an arbitrary scaling factor w.r.t. depth and translation, and (2) the lack of geometry priors for predicting complex BEV layout. Consequently, monocular depth and VO predictions need to be rescaled with a scaling ratio derived from ground truth \cite{monodepth2,bian2019unsupervised}, which is not appealing in real-world applications. And previous BEV methods \cite{mani2020monolayout,philion2020lift,yang2021projecting} usually predict the BEV layout of different semantic categories separately and ignore potentially useful geometry clues such as the depth order between cars.

In this paper, we propose to handle these three tasks simultaneously and provide complementary information for each other to address the aforementioned issues. We are inspired by the following two key observations. First, we note that the BEV road layout can provide absolute scale under the weak assumption that the road is flat, which allows us to exploit the cross-view geometric transformation and obtain a depth map with an absolute scale corresponding to the distance field that existed in the layout. As a result, the absolute scale can be introduced into our depth and VO predictions, resolving the scale ambiguity problem. Second, the learned depth predictions can provide useful priors about scene geometry (e.g., the relationship between near and far as well as overlay and underlay between objects and roads in the scene) to help solve the challenges (e.g., occlusions) in BEV layout estimation. 

To this end, we propose a novel joint perception network named JPerceiver that can estimate scale-aware depth and VO as well as BEV layout of roads and vehicles simultaneously, as shown in Fig.~\ref{fig:openimage}. JPerceiver follows the multi-task learning framework, consisting of three networks for depth, pose and layout, respectively, which can be efficiently trained in an end-to-end manner. Specifically, we design a cross-view geometric transformation-based (CGT) scale loss to propagate the absolute scale from the road layout to depth and VO. Meanwhile, a cross-view and cross-modal transfer (CCT) module is devised to leverage the depth clues for inferring the road and vehicle layouts through an attention mechanism. Our proposed scale loss and CCT module not only promote inter-task knowledge transfer but also benefit the feature learning of each task via network forward computation and gradient back-propagation. 

The contributions of this paper are summarized as follows: 1) we propose the first joint perception framework JPerceiver for depth, VO and BEV layout estimation simultaneously; 2) we design a CGT scale loss to leverage the absolute scale information from the BEV layout to achieve scare-aware depth and VO; 3) we devise a CCT module that leverages the depth clues to help reason the spatial relationships between roads and vehicles implicitly, and facilitates the feature learning for BEV layout estimation; and 4) we conduct extensive experiments on public benchmarks and show that JPerceiver outperforms the state-of-the-art methods on the above three tasks by a large margin.

\section{Related Work}

\textbf{Self-supervised depth estimation and VO.} 
SfMLearner \cite{zhou2017unsupervised} is one of the first works that propose to optimize depth and pose jointly in a self-supervised manner, utilizing the photometric consistency among continuous frames. 
Though this self-supervised learning scheme has drawn great attention from researchers and achieved promising results~\cite{wang2018learning,mahjourian2018unsupervised,monodepth2,shu2020featdepth,monodepth2}, current monocular unsupervised methods still suffer from the scale ambiguity problem. McCraith et at.~\cite{mccraith2020calibrating} fit sample points to get the road plane estimation in the 3D world during test to obtain scale hint, but the hard formulation limits its general applicability. DNet \cite{xue2020toward} proposed to recover the scale by calculating the ratio of the estimated camera height and a given one, which requires a visible ground plane to be detected during inference. Wagstaff and Kelly~\cite{wagstaff2020self} also use the camera height as the scale hint by training a plane segmentation network, which is the most similar work to ours. However, they use a three-stage training strategy to train networks separately which is much more complex than our end-to-end method. 

\textbf{BEV-based environment perception.} 
Due to the limited field of view (FOV) of FV cameras, BEV representation is commonly used in environment perception and motion planning for autonomous driving \cite{wang2021multi,reading2021categorical}. Traditional methods usually predict depth and segmentation from front images, and then warp them to BEV through inverse perspective mapping (IPM) \cite{mallot1991inverse,simond2007obstacle}, which loses a large amount of information and cause distortions due to potential occlusions. 
Recently deep learning-based methods have been developed to estimate the road and vehicle layout in the orthographic BEV plane, taking the advantage of the strong hallucination ability of CNN~\cite{lu2019monocular,mani2020monolayout,yang2021projecting,philion2020lift,fiery2021,saha2021translating}.
The newly released self-driving datasets like Argoverse \cite{Chang_2019_CVPR} and Nuscenes \cite{caesar2020nuscenes} provide a large number of BEV maps that contain annotations of drivable areas, which makes it possible to train models for BEV perception using real-world data. 
Compared with prior methods, we explore the incorporation of self-supervised depth learning explicitly, which provides an important perception output with useful geometric clues for BEV layout estimation. Dwivedi et al. \cite{dwivedi2021bird} also conduct explicit depth estimation but just take it as an intermediate process to model 3D geometry rather than a joint learning perception task. Besides, prior works usually predict different semantic categories separately, while JPerceiver exploits the synergy of different semantics and predicts the layouts of all categories simultaneously.

\textbf{MTL-based environment perception.} Recently, some multi-task learning works \cite{yin2018geonet,zou2018df,ranjan2019competitive,zhao2020collaborative,klingner2020self,chen2020puppeteergan,schon2021mgnet,Chi_2021_CVPR} propose to combine related perception tasks with depth estimation and VO to exploit complementary information such as segmentation \cite{ranjan2019competitive,klingner2020self,schon2021mgnet} and optical flow \cite{yin2018geonet,zou2018df,Chi_2021_CVPR}, which effectively boost the network performance. However none of them tackles the scale ambiguity problem of monocular depth and VO via multi-task learning, which is one of our key purposes in this paper. 
\section{Method}

\subsection{Overview of JPerciver}
\begin{figure*}[!t]
  \begin{center}
  \includegraphics[width=10.6cm]{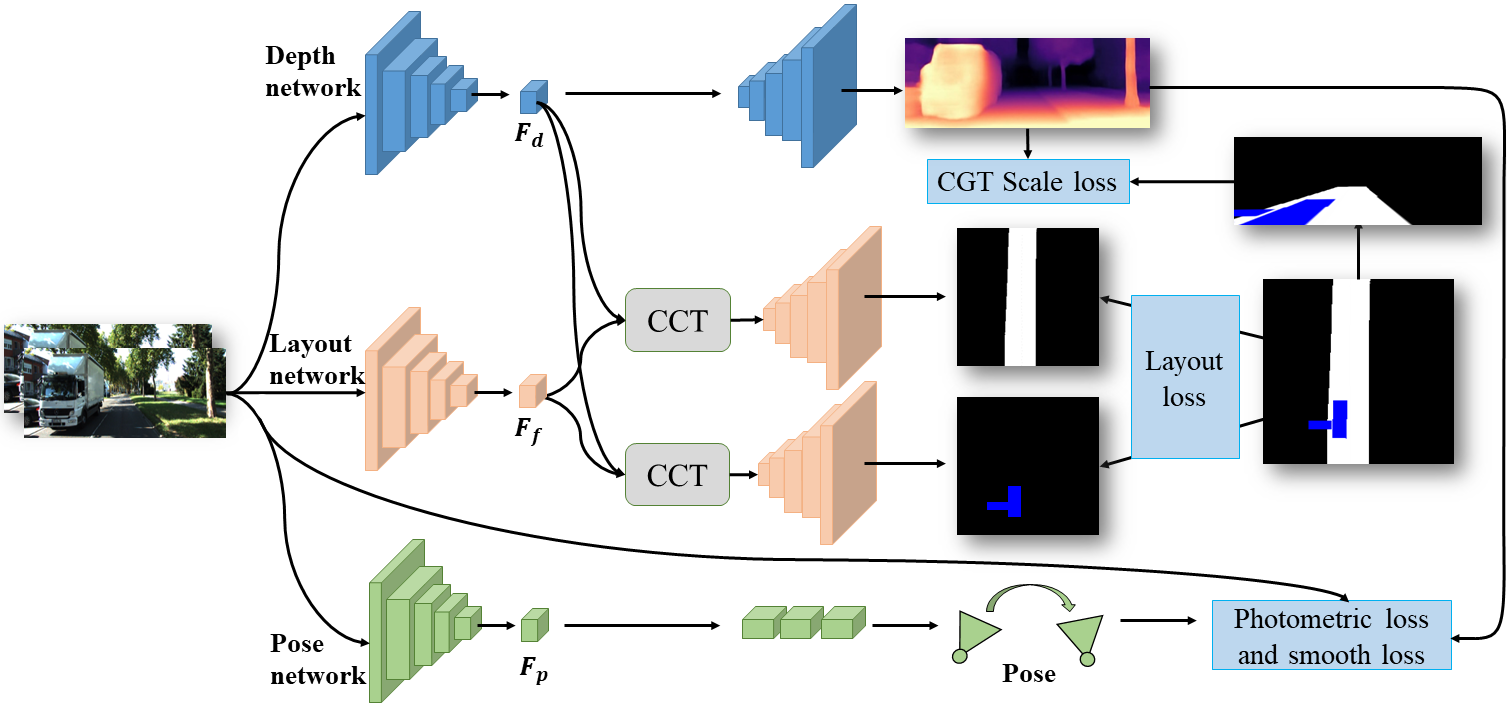} 
  \end{center}
  \caption{JPerceiver consists of three networks for depth, pose and layout estimation, and is trained in the end-to-end manner. $F_d, F_f, F_p$ represent the feature learned for three tasks and CCT denotes the cross-view and cross-modal transfer module.}
  \label{fig:overview}
\end{figure*}

As shown in Fig.~\ref{fig:overview}, JPerceiver consists of three networks for depth, pose and layout, respectively, which are all based on the encoder-decoder architecture. The depth network aims to predict the depth map $D_t$ of the current frame $I_t$, where each depth value indicates the distance between a 3D point and the camera. And the goal of the pose network is to predict the pose transformations $T_{t\rightarrow t+m}$ between the current frame and its adjacent frames $I_{t+m}$. The layout network targets to estimate the BEV layout $L_t$ of the current frame, i.e. semantic occupancy of roads and vehicles in the top-view Cartesian plane. The three networks are jointly optimized during training.
The overall objective function consists of the loss items of all the three tasks and can be formulated as:
\begin{equation}
    \ell_{total} =  \ell_{dp} + \ell_{layout},
\end{equation}
where $\ell_{dp}$ is the loss of depth and VO estimation in the self-supervised learning scheme, and $\ell_{layout}$ is the loss of the layout estimation task. We explain the details of $\ell_{dp}$ and $\ell_{layout}$ in Sec. \ref{sec3.2} and Sec. \ref{sec3.3}, respectively.

\subsection{Self-supervised Depth Estimation and VO}
\label{sec3.2}
We adopt two networks to predict depth and pose, respectively, which are jointly optimized using the photometric loss and the smoothness loss in a self-supervised manner, following the baseline method \cite{monodepth2}. We additionally devise a CGT scale loss to address the scale ambiguity problem of monocular depth and VO estimation. We describe the loss items of depth and pose networks in this section.

Self-supervised monocular depth and pose estimation is achieved by leveraging the geometry consistency among continuous frames. During training, the depth $D_t$ of current frame $I_t$ and the poses $\{T_{t\rightarrow t-1},T_{t\rightarrow t+1}\}$ between $I_t$ and its adjacent frames $\{I_{t-1},I_{t+1}\}$ are used to obtain the reconstructed current frames $\{\hat{I}_{t-1\rightarrow t},\hat{I}_{t+1\rightarrow t}\}$ via the differentiable warping function $\omega$ from $\{I_{t-1},I_{t+1}\}$:
\begin{equation}
    \hat{I}_{t+m \rightarrow t} = \omega (KT_{t\rightarrow t+m}D_tK^{-1}I_{t+m}), \ m\in\{-1, 1\}.
\end{equation}
Then, the photometric differences between $I_t$ and its reconstructed counterparts $\hat{I}_{t-1 \rightarrow t}, \hat{I}_{t+1 \rightarrow t}$ are minimized to train the depth and pose networks. We quantify the photometric differences using the SSIM and L1 losses:
\begin{equation}
    \ell_{ph} =\underset{m\in\{-1, 1\}}{min} \frac{\alpha(1-SSIM(I_t,\hat{I}_{t+m\rightarrow t}))}{2}+(1-\alpha)|I_t-\hat{I}_{t+m\rightarrow t}|,
\end{equation} 
where $\alpha$ is set to $0.85$. Following our baseline method~\cite{monodepth2}, we also take the per-pixel minimum of the photometric loss and adopt the auto-masking strategy.

To overcome the discontinuity of the predicted depth map, a smoothness loss \cite{godard2017unsupervised,monodepth2} is adopted based on the gradient of $I_t$:
\begin{equation}
    \ell_{sm} = \left|\partial_x\mu_{D_t}\right|e^{-|\partial_xI_t|}+\left|\partial_y\mu_{D_t}\right|e^{-|\partial_yI_t|},
\end{equation}
where $\mu_{D_t}$ denotes the normalized inverse depth. By minimizing the above losses, the depth network and the pose network can be optimized simultaneously.

\begin{center}
\makeatletter\def\@captype{figure}\makeatother
\begin{minipage}{.45\textwidth}

\centering
\includegraphics[height=3.5cm]{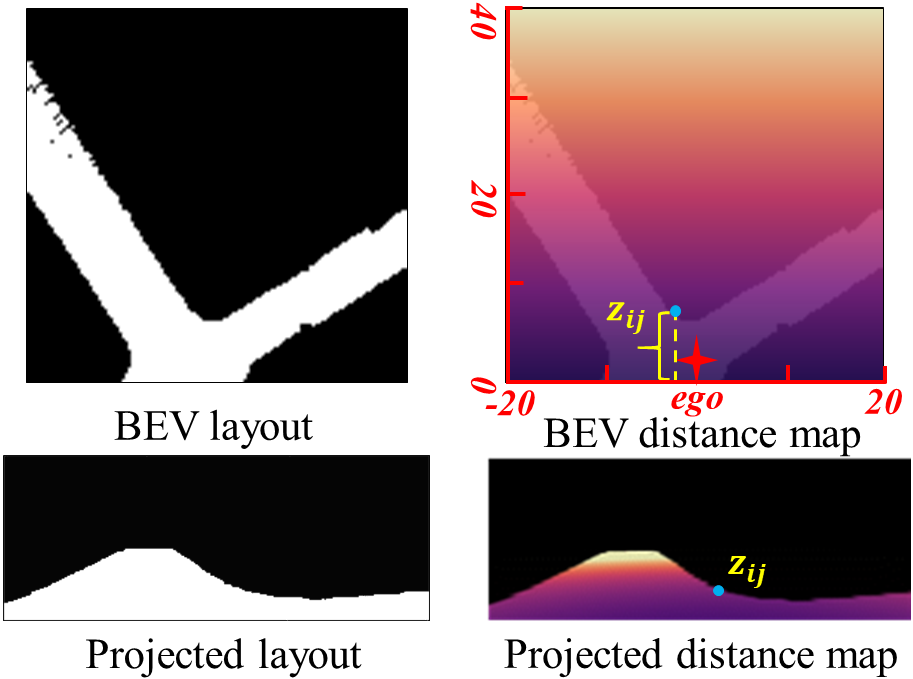}
\caption{The demonstration of CGT scale loss.}
\label{fig:scaleloss}
\end{minipage}
\makeatletter\def\@captype{figure}\makeatother
\begin{minipage}{.38\textwidth}
\centering
\includegraphics[height=3.6cm]{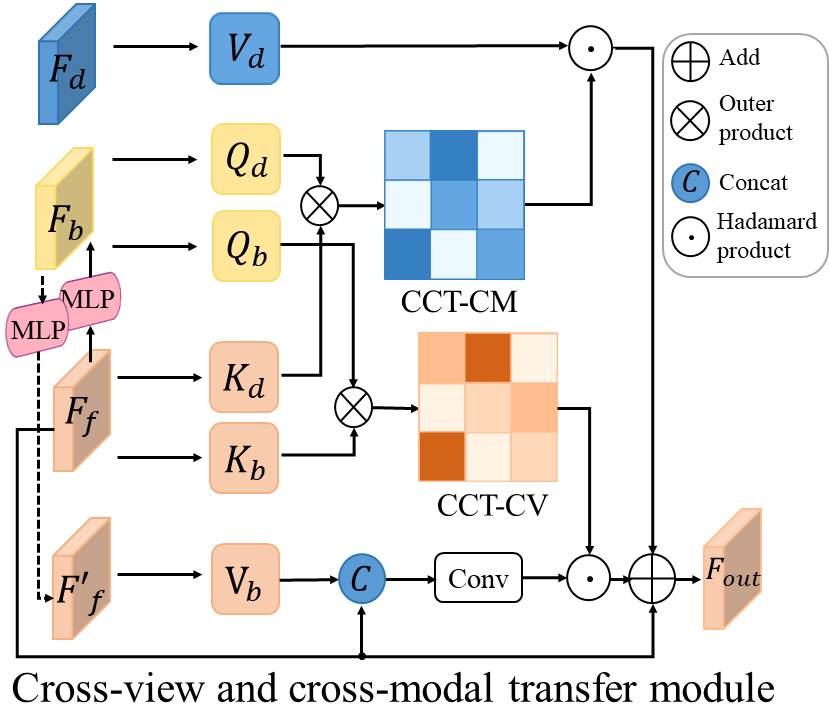}
\caption{The structure of cross-view and cross-modal transfer (CCT) module.}
\label{fig:correlation}
\end{minipage}
\end{center}

\textbf{CGT Scale Loss.}
To accomplish scale-aware environment perception, we propose the cross-view geometric transformation-based (CGT) scale loss for depth estimation and VO by utilizing the scale information in the BEV layout. Since BEV layouts demonstrate the semantic occupancy in the BEV Cartesian plane, covering the range of $Z$ meters in front of the ego vehicle and horizontally covering $\frac{Z}{2}$ meters to the left and right, respectively. It provides a natural distance field $z$ with each pixel having a metric distance value $z_{ij}$ with respect to the ego vehicle, as shown in Fig. \ref{fig:scaleloss}.
By assuming that the BEV plane is a flat plane on the ground with its origin just below the origin of the ego vehicle coordinate system, the BEV plane can then be projected to FV using the camera extrinsic parameters via a homography transformation:
\begin{equation}
    H_{bev}^{cam} = K~T_{ego}^{cam}~ T_{bev}^{ego},~~T_{ego}^{cam}~and~T_{bev}^{ego} \in \mathcal{SE}(3),
\end{equation}
where $cam, ego, bev$ represent the camera coordinate system, the ego vehicle coordinate system and the BEV ground system, respectively. $T_{bev}^{ego}$ and $T_{ego}^{cam}$ are the $\mathcal{SE}(3)$ transformations that transform the BEV plane coordinate system to the ego-vehicle coordinate system and then transform to the camera coordinate system, respectively. Therefore, the BEV distance field $z$ can be projected into FV as $z^{fv}$ as shown in Fig.~\ref{fig:scaleloss}, which is then utilized to regulate the predicted depth $d$, leading to our proposed CGT scale loss:
\begin{equation}
    \ell_{CGT} =\frac{1}{h_dw_d}\sum_{j=1}^{h_d}\sum_{i=1}^{w_d} \frac{|z^{fv}_{ij}-d_{ij}|}{z^{fv}_{ij}}, ~~if~z^{fv}_{ij}\ne 0.
\end{equation}

To learn the scale-aware depth and pose in a self-supervised manner, we take the weighted sum of $\ell_{ph}$, $\ell_{sm}$, and $\ell_{CGT}$ as the final depth-and-pose objective:
\begin{equation}
    \ell_{dp} = \ell_{ph} +  \ell_{sm} + \beta \cdot \ell_{CGT},
\end{equation}
where $\beta$ is a hyper-parameter and set to $0.1$ empirically.

\subsection{BEV Layout Estimation}
\label{sec3.3}
For layout estimation, an encoder-decoder network structure is adopted, following the prior work \cite{yang2021projecting}.
It is noteworthy that we use one shared encoder as the feature extractor and different decoders to learn BEV layouts of different semantic categories simultaneously instead of training networks for each category individually as in prior works \cite{mani2020monolayout,yang2021projecting,philion2020lift}. In addition, a CCT module is designed to strengthen feature interaction and knowledge transfer between tasks, and impose 3D geometry information for the spatial reasoning in BEV. To regularize the layout network, we combine various loss items to form a hybrid loss and achieve a balanced optimization for different categories. 

\textbf{CCT Module.}
To enhance feature interaction and impose 3D geometry information for BEV perception, we devise CCT to investigate the correlation between the FV feature $F_f$, the BEV layout feature $F_b$, the retransformed front feature $F'_f$ and the FV depth feature $F_d$, and refine the layout feature accordingly, as shown in Fig. \ref{fig:correlation}. We describe CCT in two parts, i.e. CCT-CV and CCT-CM for the cross-view module and the cross-modal module, respectively. CCT is inspired by prior work \cite{yang2021projecting}, but uses different structures and modal information. In CCT, $F_f$ and $F_d$ are extracted by the encoders of the corresponding perception branches, while $F_b$ is obtained by transforming $F_f$ to BEV with a view projection MLP, which is then re-transformed to $F'_f$ using the same MLP constrained by a cycle loss, following prior work \cite{yang2021projecting}. All the features are set to the same size, i.e., $F_f, F_d, F_b, F'_f \in \mathbb{R}^{H\times W\times D}$. In CCT-CV, a cross-attention mechanism is used to discover the geometry correspondence between FV and BEV features, which is then utilized to guide the FV information refinement and prepared for BEV reasoning. To fully exploit the FV image features, $F_b$ and $F_f$ are projected to patches ${Q_b}_i\in Q_b(i\in[1,...,HW])$ and ${K_b}_i\in K_b(i\in[1,...,HW])$, acting as the Query and Key respectively. Then each location in the FV will retrieve the correlation from every location in BEV to form a correlation matrix:
\begin{equation}
    C_b = \frac{Q_bK_b^T}{\sqrt{D}}\in \mathbb{R}^{H\times W \times H \times W},
\end{equation} where the normalization factor $\frac{1}{\sqrt{D}}$ is used to restrict the value range.
The cross-view correspondence can be identified by finding the location with the largest correlation value in $C_b$, which is differentiable by using the softmax operation: 
\begin{equation}
M_b = softmax(C_b)\in \mathbb{R}^{H\times W \times H \times W}.
\end{equation}Since $F'_f$ is obtained by first transforming the FV features to BEV and then back to FV, it contains both FV and BEV information. We thus project $F'_f$ to patches ${V_b}_i\in V_b(i\in[1,...,HW])$ and concatenate it with $F_f$ to provide the Value for the cross-view correlation after a convolution layer:
\begin{equation}
     F_{cv} =Conv( Concat(F'_f, V_b))\odot M_b.
\end{equation}

Except for utilizing the FV features, we also deploy CCT-CM to impose 3D geometry information from $F_d$. Since $F_d$ is extracted from FV images, it is reasonable to use $F_f$ as the bridge to reduce the cross-modal gap and learn the correspondence between $F_d$ and $F_b$. Thus, similar to CCT-CV, $F_b$ and $F_f$ are regarded as the Query $Q_d$ and the Key $K_d$ to calculate the correlation matrix: 
\begin{equation}
    C_d = \frac{Q_dK_d^T}{\sqrt{D}}\in \mathbb{R}^{H\times W \times H \times W}.
\end{equation} $F_d$ plays the role of Value so that we can acquire the valuable 3D geometry information correlated to the BEV information and further improve the accuracy of layout estimation. The final CCT-CM feature $F_{cm}$ is then derived as:
\begin{equation}
\begin{aligned}
    M_d &= softmax(C_d)\in \mathbb{R}^{H\times W \times H \times W},\\
 F_{cm} &= F_d \odot M_d.
\end{aligned}
\end{equation}
In the end, the original FV feature $F_f$, the cross-view correlated feature $F_{cv}$ and the cross-modal correlated feature $F_{cm}$ are summed up as the input of the layout decoder branch to conduct subsequent learning: $F_{out} = F_f + F_{cv} + F_{cm}$.

\textbf{Hybrid Loss.}
BEV layout estimation is a binary classification problem for each semantic category in the BEV grid to determine whether an area belongs to roads, cars, or backgrounds. Thus, this task is usually regularized by minimizing the difference $Diff(\cdot)$ between the predictions $L_{pred}$ and the ground truth $L_{gt}$ using Cross-Entropy (CE) or L2 loss in prior works \cite{yang2021projecting,fiery2021}:
\begin{equation}
    \ell^c_{layout} =  Diff(L_{pred}-L_{gt}), ~c \in \{c_{road}, c_{vehicle}\}.
\end{equation}
In the process of exploring our joint learning framework to predict different layouts simultaneously, we observe that a great difference exists in the characteristics and distributions of different semantic categories. For characteristics, the road layout in driving scenes usually needs to be connected, while different vehicles instead must be separated. And for distributions, more scenes with straight roads are observed than scenes with turns, which is reasonable in real-world datasets.
Such difference and imbalance increase the difficulty of BEV layout learning, especially for predicting different categories jointly, due to the failure of the simple CE or L1 losses in such circumstances. Thus, we incorporate several kinds of segmentation losses including the distribution-based CE loss, the region-based IoU loss, and the boundary loss into a hybrid loss to predict the layout for each category. First, the Weighted Binary Cross-Entropy Loss is adopted, which is most commonly used in semantic segmentation tasks:
\begin{equation}
   \ell_{WBCE} = \frac{1}{h_lw_l}\sum_{n=1}^{h_lw_l} \sum_{m=1}^M -w_m[y^m_n\cdot logx^m_n+(1-y^m_n)\cdot log(1-x^m_n)],
\end{equation} where $x_n$ and $y_n$ denote the $n$-th predicted category value and the counterpart ground truth in a layout of size $h_lw_l$, respectively. $M=2$ means whether a pixel belongs to foreground roads or cars, while $w_m$ is the hyperparameter for tackling the issue of sampling imbalance between different labels. We set $w_m$ to $5$ and $15$ for roads and vehicles respectively following \cite{yang2021projecting}.

Since the CE loss treats each pixel as an independent sample and thus neglects the interactions between nearby pixels, we then adopt the Soft IoU Loss $\ell_{IoU}$ to ensure the connectivity within the region:
\begin{equation}
    \ell_{IoU} = -\frac{1}{M} \sum_{m=1}^M\frac{\sum_{n=1}^{h_lw_l} x^m_n  y^m_n}{\sum_{n=1}^{h_lw_l} (x^m_n + y^m_n + x^m_n y^m_n)},~M=2.
\end{equation}
For the integrity of the region edges, we further use the Boundary Loss $\ell_{Bound}$ to constrain the learning of the boundary predictions, which has proven effective for mitigating issues related to regional losses in highly unbalanced segmentation problems such as medical image segmentation \cite{kervadec2019boundary,ma2020distance,xue2020shape}. It is calculated as the Hadamard product of the signed distance (SDF) map of the ground truth layout:
\begin{equation}
    \ell_{Bound} = M_{SDF}(L_{gt})\odot L_{pred},
\end{equation}
$$
M_{SDF}(L_{gt})=
\left
\{\begin{array}{rl} 
		-\inf\limits_{y\in L^b_{gt}}||x-y||_2, & x\in L^{in}_{gt},\\ 
		
		+\inf\limits_{y\in L^b_{gt}}||x-y||_2, & x\in L^{out}_{gt},\\
		0, & x\in L^b_{gt},
\end{array}\right.
$$
where $L^{in}_{gt}, L^{out}_{gt}$, and $L^b_{gt}$ represent regions inside, outside and at the foreground object boundaries in the ground truth, respectively. $||x-y||_2$ is the Euclidian distance between $x$ and $y$. 

The final loss of the layout estimation for each category then reads:
\begin{equation}
    \ell^{c_i}_{layout} =  \ell^{c_i}_{WBCE} + \lambda \cdot \ell^{c_i}_{IoU} + \lambda \cdot \ell^{c_i}_{Bound},
\end{equation} 
where $\lambda = 20$ and $c_i \in \{c_{road}, c_{vehicle}\}$.

Different from prior works \cite{yang2021projecting,philion2020lift}, our joint learning framework predict all semantic categories simultaneously instead of training a network separately for each category. The final optimization loss for our layout network reads:
\begin{equation}
    \ell_{layout} = \sum_{c_i} \ell^{c_i}_{layout}, \ c_i \in \{c_{road}, c_{vehicle}\}.
\end{equation}

\section{Experiments}
Since there is no previous work that accomplishes depth estimation, visual odometry and BEV layout estimation simultaneously, we evaluate the three tasks on their corresponding benchmarks and compared our method with the SOTA methods of each task. In addition, extensive ablation studies are performed to verify the effectiveness of our joint learning network architecture and loss items.
\subsection{Datasets.}
We evaluate our JPerceiver on three driving scene datasets, i.e., Argoverse \cite{chang2019argoverse}, Nuscenes \cite{caesar2020nuscenes} and KITTI \cite{geiger2012we}. Argoverse and Nuscenes are relatively newly published autonomous driving datasets that provide high-resolution BEV semantic occupancy labels for roads and vehicles. We evaluate the performance of BEV layout estimation on Argoverse with 6,723 training images and 2,418 validation images within the range of $40m\times40m$, and on Nuscenes with 28,130 training samples and 6,019 validation samples under two settings (in Supplementary). The ablation study for layout estimation is performed on Argoverse. Two semantic categories are included in the evaluated BEV layouts, i.e. roads and vehicles. We adopt the mean of Intersection-over-Union (mIoU) and Average Precision (mAP) as the evaluation metrics, following prior works \cite{mani2020monolayout,yang2021projecting}. Due to the insufficient annotations in KITTI, we follow prior works to use three splits for the tasks, i.e., the KITTI Odometry split (15,806 and 6,636 items for training and validation) for depth, VO, as well as road layout estimation, the KITTI Raw split (10,156 and 5,074 items for training and validation) for road layout estimation, and the KITTI 3D Object split (3,712 and 3,769 items for training and validation) for vehicle layout estimation, all within the range of $40m\times40m$. 

\noindent\textbf{Implementation Details.} We adopt the encoder-decoder structure for the three networks, all using pre-trained ResNet18 \cite{he2016deep} as the encoder backbone except that we modify the pose network to take two-frame pairs as input. Following the prior method \cite{yang2021projecting}, the input sizes of the depth and layout networks are both set to $1024 \times 1024$, while a smaller input size $192\times 640$ is used for the pose network to save computation resources since its outputs are not pixel-wise. Our model is implemented in PyTorch~\cite{paszke2017automatic} and trained for 80 epochs using Adam~\cite{kingma2014adam}, with a learning rate of $10^{-4}$ for the first 50 epochs and $10^{-5}$ for the remaining epochs. The details of the network structure are presented in Supplementary. And there is a potential promising improvement using more powerful network backbones (e.g. transformer \cite{xu2021vitae,zhang2022vitaev2}) in our method. 

\begin{table*}[!t]

\scriptsize
\centering
\caption{Quantitative comparisons (top part) and ablation study (bottom part) on Argoverse \cite{chang2019argoverse}. ``CCT-CV'' and ``CCT-CM'' denote the cross-view and cross-modal part in CCT.}
\begin{tabular}[t]{c|c|c|c|c}
\toprule
\multirow{2}*{Methods}&\multicolumn{2}{c}{Argoverse Road}&\multicolumn{2}{c}{Argoverse Vehicle}\\
 &mIoU(\%)&mAP(\%)&mIoU(\%)&mAP(\%)
  \\
  
  \hline
  VED \cite{lu2019monocular}& 72.84 & 78.11&24.16&36.83 \\
  \hline
   VPN \cite{pan2020cross}
& 71.07
 &86.83&16.58&39.73
 \\
  \hline
 Monolay \cite{mani2020monolayout}
 &73.25
 & 84.56&32.58
 & 51.06
 \\
 \hline
 PYVA \cite{yang2021projecting}
 &76.51&87.21&48.48&64.04
 \\
 \hline
 \textbf{JPeceiver(``1-1'')}
 & \textbf{77.86}
 & \textbf{90.59}&\textbf{49.94}&\textbf{65.44}\\
 \hline
\textbf{ JPerceiver(``1-2'')}
 & \textbf{77.50}
 & \textbf{90.21}&\textbf{49.45}&\textbf{65.84}\\
\hline
\hline
  Baseline(``1-1'')
 & 76.66
 & 87.17&46.97&63.36\\
 
 \hline
+CCT-CV
 & 77.76
 & 88.42&49.33&64.05\\
 \hline
  +CCT-CV+CCT-CM
 & 77.80
 & 89.00&49.39&64.86\\
 \hline
  \textbf{Ours(``1-1''+CCT+HLoss)}
 & \textbf{77.86}
 & \textbf{90.59}&\textbf{49.94}&\textbf{65.44}\\
  \hline
 Baseline(``1-2'')
 & 76.52
 & 86.54&42.77&59.39\\
 
 \hline
  +CCT-CV
 & 76.91
 & 87.19&46.46&61.02\\
 \hline
  +CCT-CV+CCT-CM
 &77.38
 &88.40&47.19&61.43\\
  \hline
  Ours(``1-2''+CCT(S)+HLoss)
 &76.81
 &89.39&48.06&63.61\\
 \hline
  \textbf{Ours(``1-2''+CCT+HLoss)}
 &\textbf{ 77.50}
 & \textbf{90.21}&\textbf{49.45}&\textbf{65.84}\\
 \bottomrule
\end{tabular}
\label{argoverse}
\end{table*}

\begin{center}
\makeatletter\def\@captype{figure}\makeatother
\begin{minipage}{.61\textwidth}
\label{fig:3}
\centering
\includegraphics[height=3.9cm]{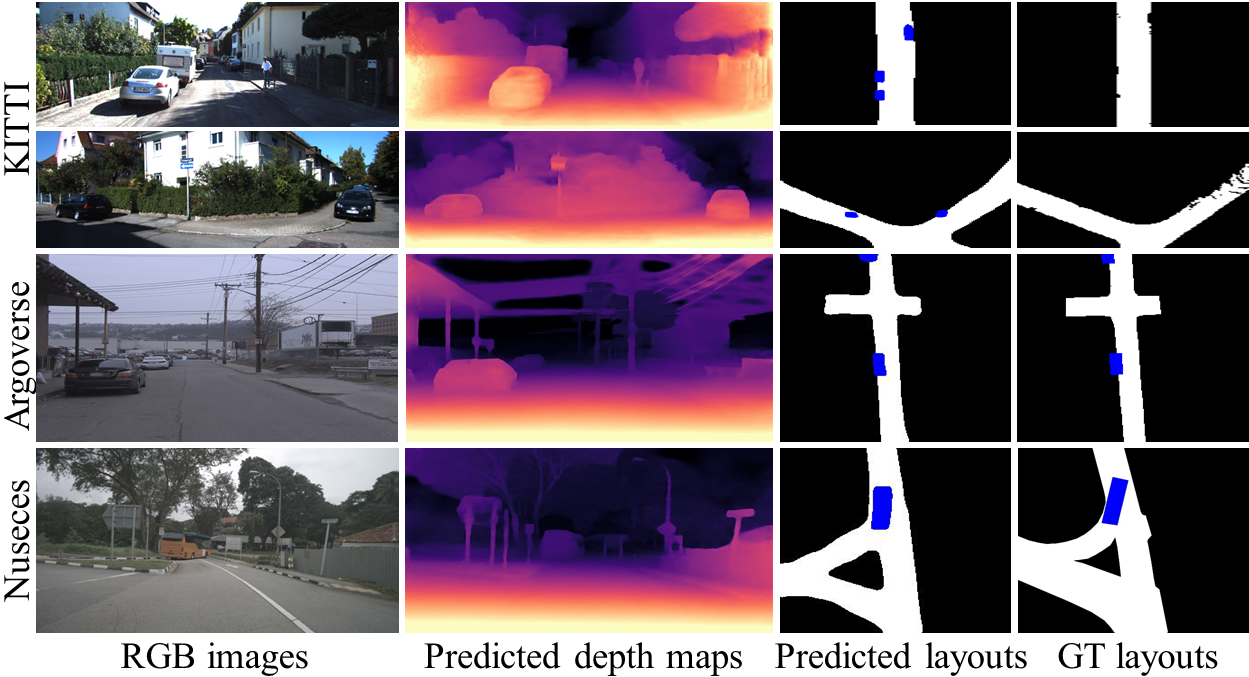}
\caption{The qualitative results of depth and layout on the three datasets. White and blue regions indicate road and vehicle layouts.}
\end{minipage}
\makeatletter\def\@captype{figure}\makeatother
\begin{minipage}{.33\textwidth}

\centering
\includegraphics[height=3.8
cm]{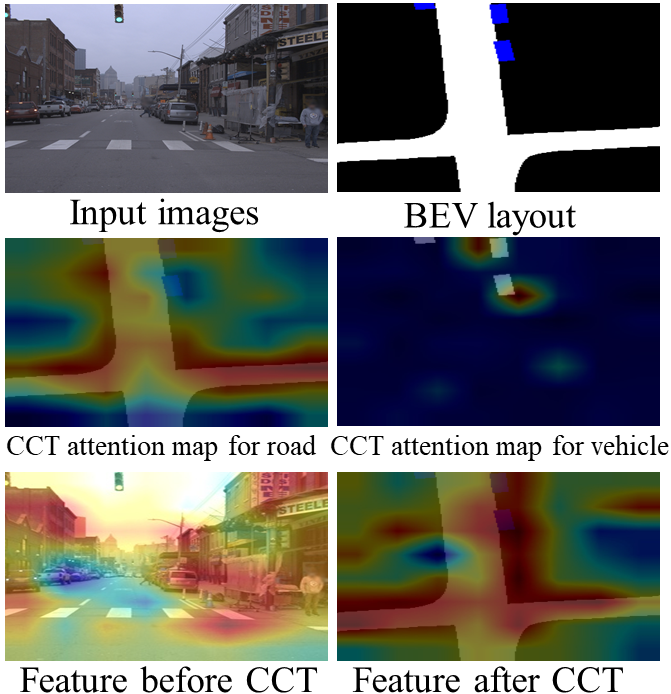}
\caption{Visualization of features and attention maps aligned with the corresponding views.}
\label{fig:attnmap}
\end{minipage}
\end{center}
\subsection{Layout Estimation}

\textbf{Argoverse.} We first quantitatively compare JPerceiver with SOTA methods on Argoverse \cite{chang2019argoverse}, including VPN \cite{pan2020cross}, Monolay \cite{mani2020monolayout} and PYVA \cite{yang2021projecting}. We compare two variants of our method, i.e., (1) JPerceiver(``1-1'') that trains two layout estimation networks separately for each category, which is also the common practice in the compared methods, and (2) JPerceiver(``1-2'') that predicts the two kinds of layouts jointly with one shared encoder. As shown in the top part of Table \ref{argoverse}, the two variants both show superiority over other approaches in the road and vehicle layout estimation. Moreover, JPerceiver(``1-2'') achieves comparable performance with JPerceiver(``1-1''), even if JPerceiver(``1-2'') only uses a shared encoder and thus is more efficient in terms of memory and computation.

\noindent\textbf{Ablation.} To investigate the effect of the component and the hybrid loss in our model, we conduct ablation studies on Argoverse and report the results in the bottom part of Table \ref{argoverse}. We take a basic ``1-1'' structure as our baseline, i.e., an encoder-decoder structure trained for each semantic category with a CE loss. We then use CCT and jointly train the depth network, pose network, as well as layout network together. Specifically, we ablate the cross-view module CCT-CV and the cross-modal module CCT-CM, respectively. As can be seen, both the cross-view and the cross-modal modules improve the performance of the baseline, and the complete CCT brings a gain of 1.14\% mIoU and 1.83\% mAP for road layout estimation and 2.42\% mIoU and 1.5\% mAP for vehicle layout estimation, respectively. After using the hybrid loss (HLoss), the performance reaches the best, i.e., 77.86\% mIoU and 90.59\% mAP for roads as well as 49.94\% mIoU and 65.44\% mAP for vehicles, respectively. We then conduct the same ablation study on the ``1-2'' structures. It is observed that using one encoder to learn representations for both two semantic layouts significantly decreases the performance of baseline models, especially for the vehicle layout, i.e., from 46.97\% mIoU and 63.36\% mAP to 42.77\% mIoU and 59.39\% mAP. After using the proposed CCT module and hybrid loss, the performance drop can be recovered, where the final model achieves comparable results as the ``1-1'' structure. We further investigate the performance of using a shared CCT module for the two semantic layouts, denoted as ``CCT(S)" in Table \ref{argoverse}. As can be seen, its results are much worse than using separate CCT for each category. It is probably due to the distinct geometry characteristics of road and vehicle, where different regions should be paid attention to as illustrated in Fig.~\ref{fig:attnmap}.

\textbf{KITTI.} We train our layout network for roads on the KITTI Odometry and the KITTI Raw splits, and the layout for vehicles on the KITTI Object split. As shown in Table \ref{bevlayout} (left), our performance on KITTI Odometry is superior to other methods w.r.t. both mIoU and mAP. The evaluation results on KITTI Raw are listed in Table \ref{bevlayout} (middle). We report our reproduced result of PYVA \cite{yang2021projecting} within parentheses because their reported results are trained with processed ground truth. Even so, our method still outperforms their reported results with a gain of 5.39\% in mAP. We observe a performance degradation from KITTI Odometry to KITTI Raw, potentially because the ground truth of the latter comes from registered semantic segmentation of Lidar scans while the former obtains the ground truth from the more accurate Semantic KITTI dataset \cite{behley2019semantickitti}, both collected by \cite{mani2020monolayout}. For vehicle layout estimation, we show the quantitative results in Table \ref{bevlayout} (right). Our method exceeds other works by a large margin, i.e., 2.06\% and 6.97\% w.r.t. mIoU and mAP.

\subsection{Depth Estimation and Visual Odometry}
KITTI presents the most commonly used dataset for depth estimation and VO. And we report our scale-aware depth and VO results on KITTI Odometry.

\textbf{Depth Estimation.} We compare with several self-supervised depth estimation methods on the KITTI Odometry test set, shown in Table \ref{depth}. The scaling factor is calculated as the average of all depth map scale ratios, which is the ratio of the median of depth values and the median of ground truth values. We use Monodepth2 \cite{monodepth2} as the baseline and DNet \cite{xue2020toward} as a representative competitor of the scale-aware methods. Though Monodepth2 \cite{monodepth2} achieves good up-to-scale accuracy, however, without the scaling factor, its performance significantly degrades. DNet \cite{xue2020toward} predicts a camera height during inference and calculates the ratio between the ground truth camera height and the predicted one to get the scaling factor. However, its output depths still need to be scaled. Differently, thanks to the CGT scale loss, our depth prediction naturally contains the absolute metric scale and does not require any scaling operation. As shown in the top part of Table \ref{depth}, our scale factor computed during inference is $1.065$ with a variance of $0.071$, while a comparable precision is also achieved.
\begin{table*}[!t]
\scriptsize
\centering
\caption{Quantitative comparisons of BEV layout estimation results on  KITTI Odometry, KITTI Raw and KITTI 3D Object.}
\setlength{\tabcolsep}{0.006\linewidth}
\begin{tabular}[t]{c|c|c|c|c|c|c}
\toprule
\multirow{2}*{Methods}&\multicolumn{2}{|c|}{KITTI Odometry Road}&\multicolumn{2}{|c|}{KITTI Raw Road}&\multicolumn{2}{|c}{KITTI 3D Object}\\
 &mIoU(\%)&mAP(\%)&mIoU(\%)&mAP(\%)&mIoU(\%)&mAP(\%)
  \\
  
  \hline
  VED \cite{lu2019monocular} & 65.74 & 67.84& 58.41 & 66.01& 20.45 & 22.5 \\
  \hline
  Mono3D \cite{chen2016monocular}&-&-&-&-& 17.11 & 26.62 \\
  \hline
  OFT \cite{roddick2018orthographic} &-&-&-&-& 25.24 & 34.69 \\
  \hline
   VPN \cite{pan2020cross}
& 66.81&81.79& 59.58& 79.07& 16.80& 35.54
 \\
  \hline
 Monolay \cite{mani2020monolayout}
  &76.81& 85.25&66.02& 75.73 &30.18&45.91
 \\
 \hline
 PYVA \cite{yang2021projecting}
 & 77.49
 & 86.69& 68.34 (65.52)  & 80.78 (79.52)& 38.79
 & 50.26
 \\
 
 \hline
 \textbf{JPeceiver}
 & \textbf{78.13}
 & \textbf{89.57}& 66.39
 &\textbf{ 86.17} & \textbf{40.85}
 & \textbf{57.23}\\
 \bottomrule
\end{tabular}
\label{bevlayout}
\end{table*}
\begin{table*}[!t]

\scriptsize

\centering
\caption{Quantitative comparisons and ablation study for depth estimation. ``w'' and ``w/o'' denote evaluation results with or without rescaling by the scale factor, which is calculated during inference.}
\setlength{\tabcolsep}{0.0015\linewidth}
\begin{tabular}[t]{c|c|c|c|c|c|c|c}
\toprule
Methods&Resolution&Scaling& Abs Rel ($\downarrow$)&Sq Rel($\downarrow$)& RMSE($\downarrow$)& RMSE log($\downarrow$)& Scale factor\\
  \hline
  \multirow{2}*{Monodepth2 \cite{monodepth2}}&\multirow{2}*{$1024 \times 1024$}
&w& \textbf{0.113}
 &0.526&3.656&0.181&$42.044 \pm 0.076$

 \\
 
&&w/o& 0.976
 &13.687&17.128&3.754
&--
 \\ 
  \hline
 \multirow{2}*{DNet \cite{xue2020toward}}& \multirow{2}*{$1024 \times 1024$} &w& 0.121&0.582& 3.762 &0.192& $34.393\pm 0.077$
\\

  &&w/o& 0.970 & 13.528&17.028&3.545 &--\\
 
\hline
  \multirow{2}*{\textbf{JPerceiver}}& \multirow{2}*{$1024 \times 1024$}
&w& 0.116
 &\textbf{0.517}&\textbf{3.573}&\textbf{0.180}&\textbf{1.065} $\pm$ \textbf{0.071}

 \\
 
&&w/o& \textbf{0.112}
 &\textbf{0.559}&\textbf{3.817}&\textbf{0.196}
&--
 \\
 \hline
 \hline
  \multirow{2}*{Baseline}& \multirow{2}*{$512 \times 512$}
&w& 0.120
 &0.550&3.670&0.184&$39.452 \pm 0.077$

 \\
 
&&w/o& 0.974
 &13.616&17.073&0.179
&--
 \\
 \hline
  \multirow{2}*{+CCT}& \multirow{2}*{$512 \times 512$}
&w& 0.108
 &0.505&3.574&0.179&$37.711 \pm 0.067$

 \\
 
&&w/o& 0.973
 &13.616&17.083&3.645
&--
 \\
 \hline
  \multirow{2}*{+scale loss}& \multirow{2}*{$512 \times 512$}
&w& 0.135
 &0.633&3.860&0.194&$1.088 \pm 0.093$

 \\
 
&&w/o& 0.125
 &0.643&4.092&0.211
&--
 \\
 \hline
  \multirow{2}*{Ours}& \multirow{2}*{$512 \times 512$}
&w& 0.128
 &0.574&3.739&0.189&$1.099 \pm 0.085$

 \\
 
&&w/o& 0.122
 &0.628&3.952&0.205
&--
 \\
 \bottomrule

\end{tabular}

\label{depth}
\end{table*}

\noindent \textbf{Ablation.}
We further conduct the ablation study for depth and report the results in the bottom part of Table \ref{depth} with the input resolution of $512 \times 512$. Introducing CCT on the baseline structure boosts the depth estimation results no matter with or without scaling, and the scaling factor is similar to the baseline counterpart. We then add the CGT scale loss to the baseline to validate its feasibility, resulting in a nearly perfect scaling ratio. However, since our CGT scale loss only takes regional pixels in ground areas into account instead of using all pixels and is based on an assumption that the ground plane is flat, a less accurate result is observed for the overall prediction. Our full model achieves comparable depth performance with the baseline but up to a metric scale. Of note is that all variants with our scale loss obtain lower Abs Rel error without scaling while worse results in other metrics compared with the results with scaling. It may be because the calculation of our Scale loss is the same as the Abs Rel metric. 
Different ways of calculating and utilizing the scale loss on other baselines might be explored to further improve the estimation accuracy in future work.

\textbf{Visual Odometry.}
We train Perceiver on the KITTI Odometry sequences 01-06 and 08-09, and use the sequences 07 and 10 as our test set for evaluating our model for VO.
We compare with several self-supervised visual odometry methods in Table \ref{tab:710}, including SfMLearner \cite{zhou2017unsupervised}, GeoNet \cite{yin2018geonet}, Monodepth2 \cite{monodepth2}, SC-Sfmlearner \cite{bian2019unsupervised}, which are trained on sequence 00-08. ``Scaling'' means the scaling method is used during inference. ``GT'' means the scaling factor for correcting the predictions comes from the ground truth. Dnet \cite{xue2020toward}, LSR \cite{wagstaff2020self} and Ours all borrow information from the road plane to recover the scale but in different ways. Of note is that Dnet \cite{xue2020toward} needs invisible ground plane to predict camera height during inference for scale recovery, which is not required by our method. While LSR~\cite{wagstaff2020self} incorporates a front-view ground plane estimation task, their networks are trained in a serial way, i.e. unscaled depth and VO network, plane segmentation network, and then scaled depth and VO network. In comparison, our method can produce scaled VO results during inference without any hint by using our CGT scale loss. In addition, our method is superior to other competitors w.r.t. the average translational and rotational RMSE drift metrics. The comparison of VO trajectories with other methods without rescaling is shown in Fig. \ref{fig:VO}, which further proves the effectiveness of our scale loss. 

\makeatletter\def\@captype{figure}\makeatother
\begin{minipage}{.33\textwidth}
\centering
\includegraphics[height=3.4cm]{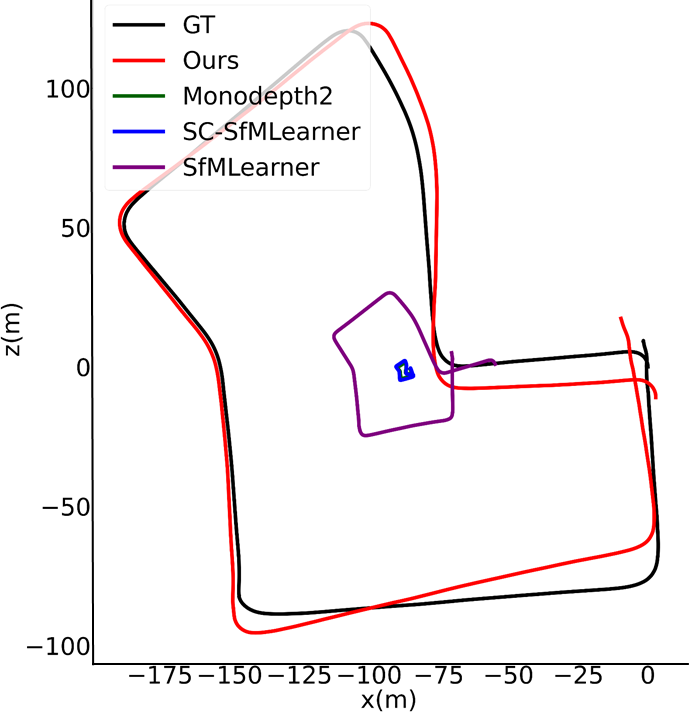}
\caption{The comparison of VO trajectories on sequence 07.}
\label{fig:VO}
\end{minipage}
\makeatletter\def\@captype{table}\makeatother
\begin{minipage}{.63\textwidth}
\scriptsize

\centering
\caption{The comparison of Visual Odometry. $t_{err}$ is the average translational RMSE drift $(\%)$ on length from 100, 200 to 800 m, and $r_{err}$ is average rotational RMSE drift $(^{\circ}/100m)$ on length from 100, 200 to 800 m.}
\begin{tabular}[t]{|c|c|c|c|c|c|c|}
\hline
\multirow{2}*{Methods}&\multirow{2}*{Scaling}&\multicolumn{2}{c|}{Sequence 07}&\multicolumn{2}{c|}{Sequence 10}\\
 &&$t_{err}$&$r_{err}$&$t_{err}$&$r_{err}$
  \\
  
  \hline
  SfMLearner \cite{zhou2017unsupervised}& GT& 12.61 & 6.31&15.25&4.06 \\
  \hline
  GeoNet \cite{yin2018geonet}
&GT& 8.27
 &5.93&20.73&9.04
 \\
  \hline
 Monodepth2 \cite{monodepth2}
 &GT&8.85
 & 5.32&11.60
 & 5.72
 \\
 \hline
 SC-Sfmlearner \cite{bian2019unsupervised}
 &GT&8.29&4.53&10.74&4.58
 \\
 \hline
 Dnet \cite{xue2020toward}
&Camera height & -
 & -&13.98&4.07\\
 \hline
 LSR \cite{wagstaff2020self}
&None & -
 & -&10.54&4.03\\
 \hline
 \textbf{JPerveiver}
&None & \textbf{4.57}
 & \textbf{2.94}&\textbf{7.52}&\textbf{3.83}\\
 \hline
\end{tabular}

\label{tab:710}
\end{minipage}

\subsection{Model Complexity}
We compare the complexity of our JPerciever with the single-task competitors and the simply combined model in Table \ref{complexity}. Since our method can predict the depth, VO, and BEV layouts of the two semantic categories simultaneously, it is less complex in terms of parameters (M), computations (FLOPs), and inference speed (FPS) compared with the combined model while producing better prediction in all the three tasks. Besides, our JPerciever benefits more from parallel acceleration as shown in the last column, where a batch size of 6 is used.

\begin{table*}[t]

\scriptsize
\centering

\begin{tabular}[t]{|c|c|c|c|c|c|}
\hline
Methods&Task&Params(M)& Flops(G)&FPS(BS=1)&FPS(BS=6)\\
  
  \hline
 
  Monodepth2 \cite{monodepth2}&depth \& pose&
39.33& 26.93
 & 30.3&36.9 \\
 \hline
PYVA \cite{yang2021projecting}&layout& 29.73& 20.42
 &65.2&72 \\

 \hline
Monodepth2 \cite{monodepth2}
&\multirow{2}*{~depth \& pose \& layout~}
&\multirow{2}*{69.06}& \multirow{2}*{47.35}
 &\multirow{2}*{15.6}&\multirow{2}*{18.6}
 \\
 +$2 \times$PYVA \cite{yang2021projecting}&
&& 
 &&

 \\
\hline
 JPerceiver
&~depth \& pose \& layout~&57.15& 37.69
 &19.8& 26.9

 \\
 \hline
\end{tabular}
\caption{The Analysis of model complexity with input resolution $512 \times 512$ using one single GPU.} 
\label{complexity}
\end{table*}

\section{Conclusion}
In this paper, we propose a joint perception framework named JPerceiver for the autonomous driving scenarios, which accomplishes scale-aware depth estimation, visual odometry, and also BEV layout estimation of multiple semantic categories simultaneously. 
To realize the joint learning of multiple tasks, we introduce a cross-view and cross-modal transfer module and fully make use of the metric scale from the BEV layout to devise a cross-view geometry transformation-based scale loss to obtain scale-aware predictions. Our method achieves better performance towards all the above three perception tasks using less computation resource and training time. We hope our work can provide valuable insight to the future study of designing more effective joint environment perception model.\\

\section*{JPerceiver: Joint Perception Network for Depth, Pose and Layout Estimation in Driving Scenes (Supplementary Material)}
In our supplementary material, we first show more additional quantitative and qualitative evaluation results (Sec. \ref{evalresults}), towards layout estimation (Sec. \ref{layout}), depth estimation (Sec. \ref{depth}) and visual odometry (Sec. \ref{VO}). Besides, we supplement the ablation study of the network architecture designing (Sec. \ref{networkalation}) and the implementation details (Sec. \ref{netwokdetail}).
\section{Additional Evaluation Results}
\label{evalresults}
\subsection{Layout Estimation}
\label{layout}
\textbf{BEV layout estimation on Nuscenes.}
In nuscenes\cite{caesar2020nuscenes}, we compare with more recent methods that either take six-camera~\cite{pan2020cross,saha2021enabling,philion2020lift,fiery2021} or one-camera~\cite{lu2019monocular,yang2021projecting} images as input, and evaluate it under two settings, i.e., predicting a BEV layout in a range of $50m \times 50m$ (denoted as Setting 1 in Table~\ref{nuscenes}) and $100m \times 100m$ (denoted as Setting 2 in Table~\ref{nuscenes}). It is noteworthy that we choose to predict in an area of the same size for fairness, which covers the range of $Z$ m in front of the ego vehicle and horizontally covers $\frac{Z}{2}$ m to the left and right. The results of six-camera methods are retrained to predict only two classes (drivable area and car). And the results of PYVA \cite{yang2021projecting} are also trained using their provided codes on the Nuscenes dataset. 
According to Table~\ref{nuscenes}, while our method takes only one-camera images as input, we have achieved comparable or superior results to the SOTA methods \cite{saha2021enabling,pan2020cross} that take six-camera input under Setting 1. In addition, our method also surpasses the other methods that take one-camera input in both settings by a large margin.

\textbf{Ablation of different losses.}
We conduct an ablation study of different components of Hybrid Loss on the KITTI Object dataset. As shown in Table \ref{tab:lossablation}, Hybrid loss achieves superior performance w.r.t. both mIoU and mAP, consisting of CE, IoU and Boundary Losses.

\begin{table*}[!htp]

\scriptsize

\centering
\scriptsize
\begin{tabular}[t]{|c|c|c|c|c|c|c|c|c|c|}

  \hline
\multirow{3}*{Methods}&\multirow{3}*{Input}&\multicolumn{4}{c|}{Nuscenes Road}&\multicolumn{4}{c|}{Nuscenes Vehicle}\\
&&\multicolumn{2}{c|}{Setting 1}&\multicolumn{2}{c|}{Setting 2}&\multicolumn{2}{c|}{Setting 1}&\multicolumn{2}{c|}{Setting 2}\\

 &&mIoU(\%) &mAP(\%) &mIoU(\%) &mAP(\%) &mIoU(\%) &mAP(\%) &mIoU(\%) &mAP(\%) 
  \\
  
  \hline
  VED\cite{lu2019monocular} &1c& 63.8 & -&-&- & 15.6 & -&-&-\\
  \hline
  PON\cite{roddick2020predicting} &6c& 70.5 & -&-&- & 27.6 & -&-&-\\
  \hline
   VPN\cite{pan2020cross}
&6c6d& 69.4
 &-&-&-& 28.3
 &-&-&-
 \\
 \hline
 Lift-Splat\cite{philion2020lift}
&6c& -
 &-&72.94&-& -
 &-&32.1&-
 \\
 \hline
 Fiery\cite{fiery2021}
&6c& -
 &-&-&-& 37.7
 &-&35.8&-
 \\
 \hline
 PYVA\cite{yang2021projecting}&1c
 &77.09&86.19&66.55&80.42& 24.34
 &39.96&20.15&29.29
 \\
 \hline
\textbf{JPerceiver}&1c
 &\textbf{79.02}
 
 &\textbf{90.73} &68.54&\textbf{84.73}&33.01&\textbf{49.85}&24.90&\textbf{41.12}\\
 \hline

\end{tabular}

\caption{Quantitative comparisons on Nuscenes \cite{caesar2020nuscenes}. ``c'' and ``d'' in the Input column denote camera images and depth maps, e.g. ``6c6d'' means six RGB images and six depth maps are taken as input.}
\label{nuscenes}
\end{table*}
\begin{table*}[htbp]
\scriptsize
    \centering
    \begin{tabular}[t]{|c|c|c|}
  \hline

\multicolumn{3}{|c|}{KITTI 3D Object}\\
 Loss items&mIoU(\%)&mAP(\%)
  \\
  
  \hline
  CE & 39.45 & 53.89 \\
  \hline
  IoU& 41.11 & 55.80 \\
  \hline
   Boundary
& 36.08
 & 60.2
 \\
  \hline
  CE+IoU & 40.46 & 56.62 \\
  \hline
 CE+Boundary
 &39.71
 &57.47
 \\
 \hline
 IoU+Boundary
 & 40.33
 & 56.35
 \\
 \hline
 CE+IoU+Boundary
 & 40.85
 & 57.23\\
 \hline
\end{tabular}
    \caption{The experiment results of ablation study of different losses used for layout estimation}
    \label{tab:lossablation}
\end{table*}

\begin{figure*}[!t]
 \begin{center}
  \includegraphics[width=12.4cm]{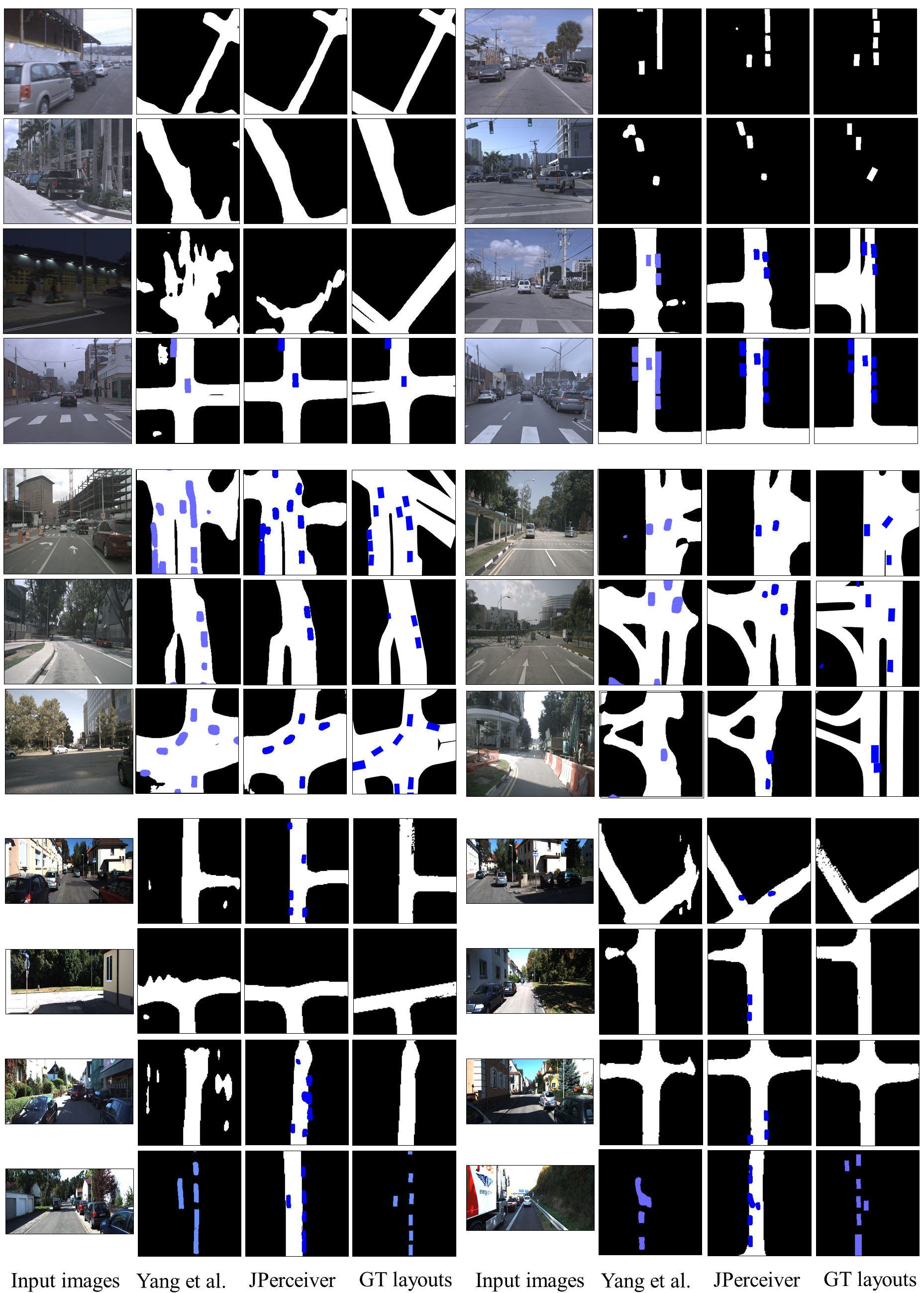} 
 \end{center}
 \caption{Visualization examples of the road and vehicle BEV layouts on Argoverse \cite{chang2019argoverse}, Nuscenes \cite{caesar2020nuscenes} and KITTI \cite{geiger2012we}, compared with the SOTA method \cite{yang2021projecting}. In the KITTI dataset, the Odometry and Raw split only provide road layout label while the Object split only provide vehicle layout label. However, our JPerceiver can predict the layouts of roads and vehicles simultaneously after training different branches on different splits, which demonstrates its superior ability of generalization in unseen scenarios. More visualization results are shown in the supplemental videos.}
\label{fig:layouts}
\end{figure*}

\textbf{Qualitative results.} As shown in Fig. \ref{fig:layouts}, we compare the estimated BEV layouts with the state-of-the-art (SOTA) method and corresponding ground truth in various test scenarios of different datasets, including Argoverse \cite{chang2019argoverse} (top), Nuscenes \cite{caesar2020nuscenes} (middle) and KITTI \cite{geiger2012we} (bottom). Notably, the estimation results of different semantic categories, i.e. road and vehicle layout are estimated simultaneously in our JPerceiver, while manually overlayed from twice inference results for the prior method \cite{yang2021projecting}.
\subsection{Depth Estimation}
\label{depth}
\begin{figure*}[!t]
 \begin{center}
  \includegraphics[width=12.4cm]{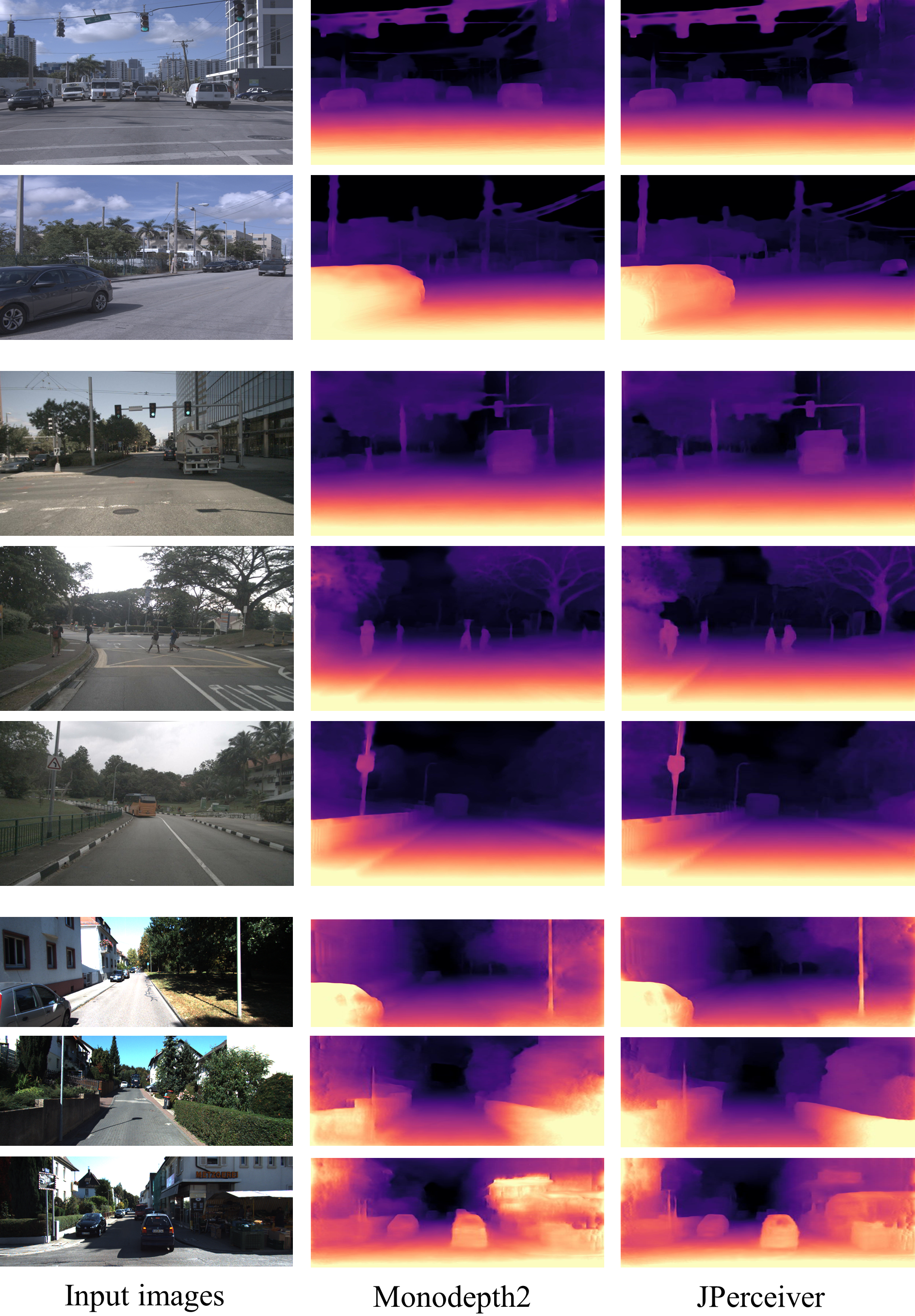} 
 \end{center}
 \caption{Visualization of predicted depth maps and qualitative comparison with our baseline method \cite{monodepth2} on the Argoverse \cite{chang2019argoverse} (top part), Nuscenes \cite{caesar2020nuscenes} (middle part) and KITTI \cite{geiger2012we} (bottom part) datasets.}
\label{fig:suppdepth}
\end{figure*}

In Fig. \ref{fig:suppdepth}, we demonstrate more visualization results of depth estimation and the comparison with our baseline method Monodepth2 \cite{monodepth2} on the validation or test sets of three datasets, i.e. Argoverse \cite{chang2019argoverse}, Nuscenes \cite{caesar2020nuscenes} and KITTI \cite{geiger2012we}.

\begin{figure}
\begin{center}
\begin{minipage}{1\textwidth}
\centering
\includegraphics[width=0.7\textwidth]{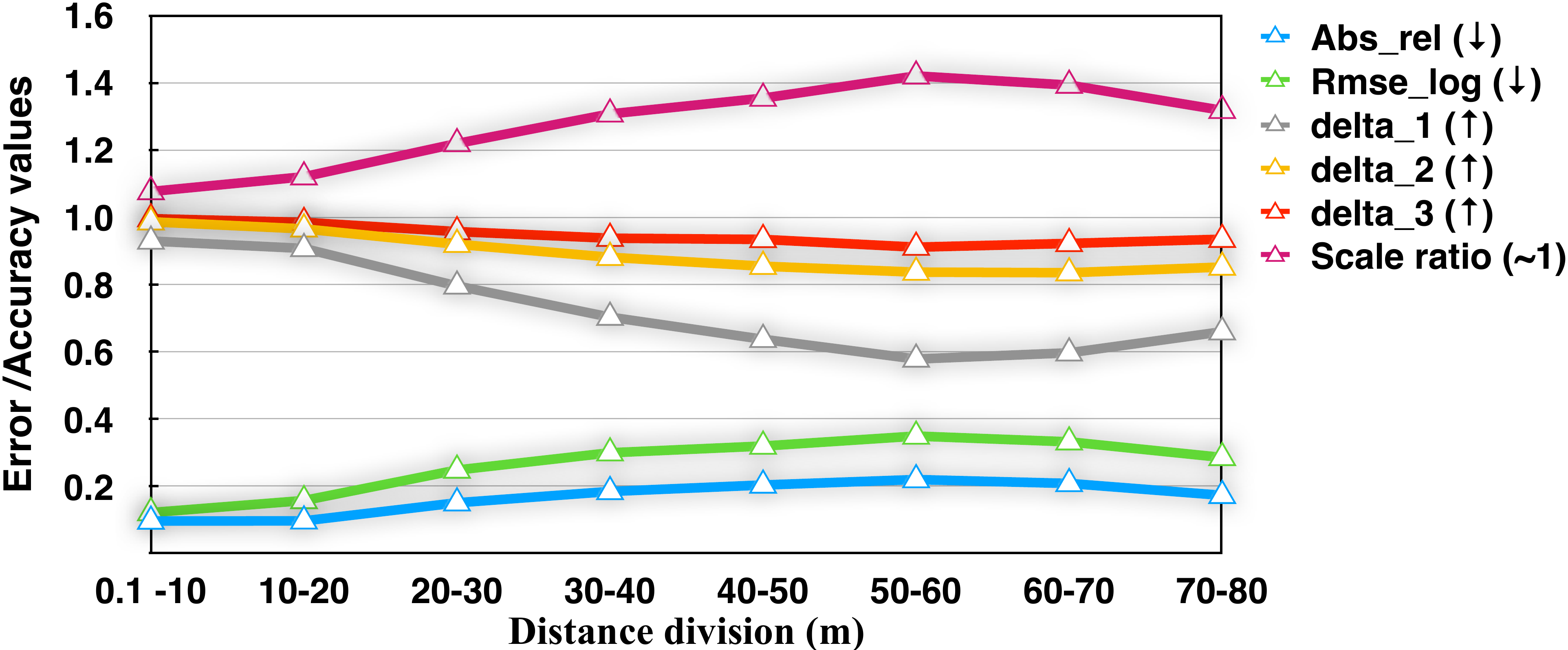}
\end{minipage}
\end{center}

\caption{The relevance between the depth metrics and distances. }
\label{errormap}

\end{figure}

\textbf{Analysis of CGT loss and depth estimation.} 
The CGT scale loss does not use all pixels in the road plane but chooses the region with vehicle occupancy removed to impose scale constraint. Due to the flat-ground assumption and the few-pixel constraint, CGT does not act as strict supervision but only provides scale at a limited range. Thus, we analyze the relevance between depth metrics and distances. As shown in Fig. \ref{errormap}, the metrics and scale factor fluctuate with regard to distance but just slightly. This is expected since long-distance depth estimation is more difficult than that in close range.

\subsection{Visual Odometry}
\label{VO}
\begin{figure*}[!t]
 \begin{center}
  \includegraphics[width=12.4cm]{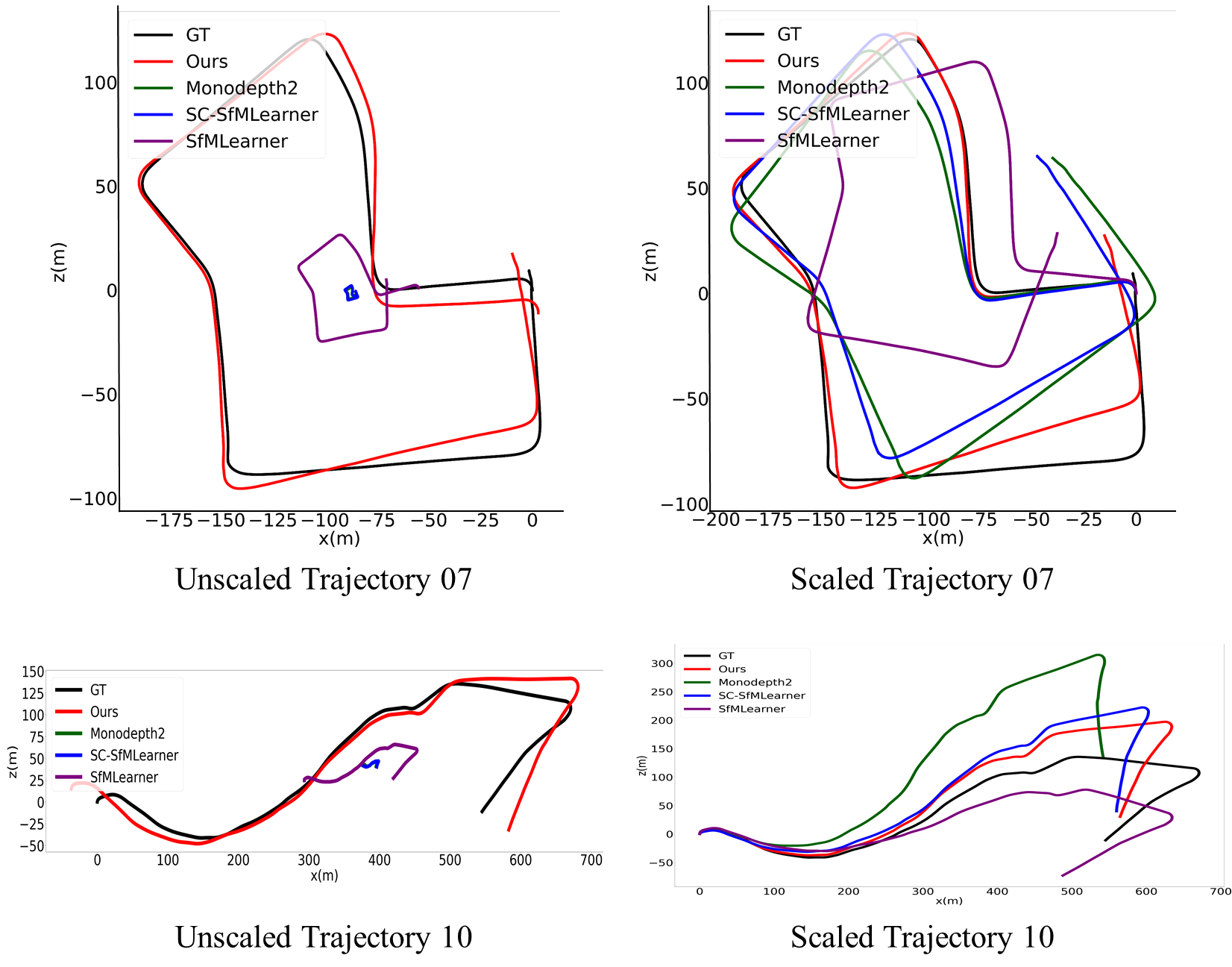} 
 \end{center}
 \caption{We demonstrate the visual odometry trajectory before and after scaling on KITTI Odometry sequences 07 and 10. More visualization results are shown in the supplemental videos.}
\label{fig:suppvo}
\end{figure*}

The visual odometry trajectories on KITTI test sets 07 and 10 are visualized in Fig. \ref{fig:suppvo}. The left part shows the trajectories without scaling, which demonstrate the absolute scale can be learned in our method. Scaled trajectories using the scaling ratio obtained from ground truth are listed in the right part, showing our predicted trajectory is much closer to the ground truth.

\makeatletter\def\@captype{table}\makeatother
\begin{minipage}{0.9\textwidth}
\scriptsize

\centering
\caption{The comparison of Visual Odometry. $t_{err}$ is the average translational RMSE drift $(\%)$ on length from 100, 200 to 800 m, and $r_{err}$ is average rotational RMSE drift $(
^{\circ}/100m)$ on length from 100, 200 to 800 m.}
\begin{tabular}[t]{|c|c|c|c|c|c|c|}

    \hline
\multirow{2}*{Methods}&\multirow{2}*{Scaling}&\multicolumn{2}{c|}{Sequence 09}&\multicolumn{2}{c|}{Sequence 10}\\
 &&$t_{err}$&$r_{err}$&$t_{err}$&$r_{err}$
  \\
  
  \hline
  SfMLearner\cite{zhou2017unsupervised}& GT& 11.32 & 4.07&15.25&4.06 \\
  \hline
  GeoNet\cite{yin2018geonet}
&GT& 28.72
 &9.8&20.73&9.04
 \\
  \hline
 Monodepth2\cite{monodepth2}
 &GT&11.47
 & 3.2&11.60
 & 5.72
 \\
 \hline
 SC-Sfmlearner\cite{bian2019unsupervised}
 &GT&7.64&2.19&10.74&4.58
 \\
 \hline
 Dnet\cite{xue2020toward}
&Camera height & 7.23
 & 1.91&13.98&4.07\\
 \hline
 LSR\cite{wagstaff2020self}
&None & \textbf{5.93}
 & 1.67&10.54&4.03\\
 \hline
 \textbf{JPerveiver}
&None & 6.81
 & \textbf{1.18}&\textbf{6.92}&\textbf{1.47}\\
 \hline
\end{tabular}

\label{tab:710supp}
\end{minipage}
\vspace{0.6cm}

\textbf{Comparison with more methods on test sequences 09 and 10.} Following the commonly used protocol in self-supervised depth estimation and visual odometry, we retrain our models using sequences 00-08 as training set and 09-10 as the test set. For the data without layout labels, we use the models pretrained on those sequences with the layout labels to generate pseudo labels, which is demonstrated feasible to provide absolute scale for self-supervised depth estimation and visual odometry. Due to the promising ability of generalization, the pretrained model can be used to generate labels for more data set to help complete scale-aware perception in future work. The quantitative comparison of self-supervised visual odometry is shown in Table \ref{tab:710supp}.

\section{Network Architecture}
\label{network}
\subsection{Ablation Study for Network Architecture}
\label{networkalation}
To explore the effectiveness of the joint learning architecture, we complete an additional ablation study via using one encoder as the feature extractor for depth network and layout network on the KITTI Odometry dataset. As shown in Table \ref{suppodometry} for layout estimation, Table \ref{suppdepthcamparison} for depth estimation and Table \ref{suppvotable} for visual odometry, the effectiveness of all three tasks have deteriorated using a shared encoder, especially for depth estimation and visual odometry.

\makeatletter\def\@captype{table}\makeatother
\begin{minipage}{0.4\linewidth}
\scriptsize
\centering
\ttabbox{\caption{Road layout estimation results on KITTI Odometry. ``$E_S$'' denotes the variant using shared encoder.}}{%
\label{suppodometry}
\begin{tabular}[t]{|c|c|c|}
  \hline
\multicolumn{3}{|c|}{KITTI Odometry Road}\\
 Methods&mIoU(\%)&mAP(\%)
  \\
  
  \hline
\textbf{JPerceiver}
 & \textbf{78.13}
 & \textbf{89.57}\\
 \hline
 JPerceiver$-E_S$&77.53&88.16\\
 \hline
\end{tabular}}
\end{minipage}
\makeatletter\def\@captype{table}\makeatother
\begin{minipage}{0.45\linewidth}
\scriptsize
\centering
\ttabbox{\caption{The comparison of Visual Odometry. $t_{err}$ is the average translational RMSE drift $(\%)$ on length from 100, 200 to 800 m, and $r_{err}$ is average rotational RMSE drift (
$ ^{\circ}/100m)$ on length from 100, 200 to 800 m. ``$E_S$'' denotes the variant using shared encoder.}}{%

\label{suppvotable}
\begin{tabular}[t]{|c|c|c|c|c|c|c|}

  \hline
\multirow{2}*{Methods}&\multirow{2}*{Scaling}&\multicolumn{2}{c|}{Sequence 07}&\multicolumn{2}{c|}{Sequence 10}\\
 &&$t_{err}$&$r_{err}$&$t_{err}$&$r_{err}$
  \\

 \hline
 \textbf{JPerveiver}
&None & \textbf{4.57}
 & \textbf{2.94}&\textbf{7.52}&\textbf{3.83}\\
 \hline
 JPerveiver$-E_S$& None & 9.73&5.72&15.75&6.9\\
 \hline
\end{tabular}}

\label{tab:supp710}
\end{minipage}
\subsection{Network Details}
\label{netwokdetail}
\textbf{Input and Output.} We take in the RGB images of size $1024\times 1024$ as input and output depth map, BEV layout and poses simultaneously. For layout network, the estimated BEV layouts of size $256\times 256$ represent a specific region in the BEV plane, such as $40m \times 40m$ in Argoverse \cite{chang2019argoverse} and KITTI \cite{geiger2012we}, and $50m\times 50m$ or $100m \times 100m$ in Nuscenes \cite{caesar2020nuscenes}.

\textbf{Encoder.} We take ResNet-18  \cite{he2016deep} as the backbone of our feature extractor for three tasks. Following the baseline method \cite{monodepth2}, we start training with weights pretrained on ImageNet \cite{russakovsky2015imagenet}. The encoder of the pose network is modified to take two-frame pair as input.

\textbf{Task-specific Decoder.} The decoder of depth network is similar to \cite{monodepth2}, using sigmoid activation functions in multi-scale side outputs and ELU nonlinear functions \cite{clevert2015fast} otherwise. While the decoder of the pose network consists of three convolution layers to predict a 6-DoF relative pose. The decoder of the layout network is composed of four deconvolution blocks to upsample the feature maps and decrease the number of feature channels, which finally arrive at the size of $256\times256\times2$ and then processed by a non-linear layer to obtain the layout.



\begin{table*}[t]

\scriptsize

\centering
\caption{Ablation study of network architecture for depth estimation. ``Scale factor'' is calculated during inference. ``w'' and ``w/o'' denote evaluation results with or without rescaling by the scale factor. ``$E_S$'' denotes the variant using shared encoder.}
\begin{tabular}[t]{c|c|c|c|c|c|c}

  \toprule
Methods&Scaling& Abs Rel ($\downarrow$)&Sq Rel($\downarrow$)& RMSE($\downarrow$)& RMSE log($\downarrow$)& Scale factor\\

\hline
  \multirow{2}*{\textbf{JPerceiver}}
&w& 0.116
 &\textbf{0.517}&\textbf{3.573}&\textbf{0.180}&1.065 $\pm$ 0.071
 \\
 
&w/o& \textbf{0.112}
 &\textbf{0.559}&\textbf{3.817}&\textbf{0.196}
&--

 \\
 \hline
 Jperceiver
&w& 0.133
 &0.646&3.853&0.195&0.990 $\pm$ 0.082

 \\
 
$-E_S$&w/o&  0.137
 &0.666&3.867&0.200
&--

 \\
 \bottomrule

\end{tabular}

\label{suppdepthcamparison}
\end{table*}
\clearpage
%
%
\bibliographystyle{splncs04}
\bibliography{eccv2022_arxiv.bbl}

\begin{thebibliography}{10}
\providecommand{\url}[1]{\texttt{#1}}
\providecommand{\urlprefix}{URL }
\providecommand{\doi}[1]{https://doi.org/#1}

\bibitem{saha2021translating}
Translating images into maps. arXiv preprint arXiv:2110.00966  (2021)

\bibitem{Badki_2021_CVPR}
Badki, A., Gallo, O., Kautz, J., Sen, P.: Binary ttc: A temporal geofence for
  autonomous navigation. In: Proceedings of the IEEE/CVF Conference on Computer
  Vision and Pattern Recognition (CVPR). pp. 12946--12955 (June 2021)

\bibitem{behley2019semantickitti}
Behley, J., Garbade, M., Milioto, A., Quenzel, J., Behnke, S., Stachniss, C.,
  Gall, J.: Semantickitti: A dataset for semantic scene understanding of lidar
  sequences. In: Proceedings of the IEEE/CVF International Conference on
  Computer Vision. pp. 9297--9307 (2019)

\bibitem{bian2019unsupervised}
Bian, J.W., Li, Z., Wang, N., Zhan, H., Shen, C., Cheng, M.M., Reid, I.:
  Unsupervised scale-consistent depth and ego-motion learning from monocular
  video. Advances in Neural Information Processing Systems  (2019)

\bibitem{caesar2020nuscenes}
Caesar, H., Bankiti, V., Lang, A.H., Vora, S., Liong, V.E., Xu, Q., Krishnan,
  A., Pan, Y., Baldan, G., Beijbom, O.: nuscenes: A multimodal dataset for
  autonomous driving. In: Proceedings of the IEEE/CVF conference on computer
  vision and pattern recognition. pp. 11621--11631 (2020)

\bibitem{Casas_2021_CVPR}
Casas, S., Sadat, A., Urtasun, R.: Mp3: A unified model to map, perceive,
  predict and plan. In: Proceedings of the IEEE/CVF Conference on Computer
  Vision and Pattern Recognition (CVPR). pp. 14403--14412 (June 2021)

\bibitem{Chang_2019_CVPR}
Chang, M.F., Lambert, J., Sangkloy, P., Singh, J., Bak, S., Hartnett, A., Wang,
  D., Carr, P., Lucey, S., Ramanan, D., Hays, J.: Argoverse: 3d tracking and
  forecasting with rich maps. In: Proceedings of the IEEE/CVF Conference on
  Computer Vision and Pattern Recognition (CVPR) (June 2019)

\bibitem{chang2019argoverse}
Chang, M.F., Lambert, J., Sangkloy, P., Singh, J., Bak, S., Hartnett, A., Wang,
  D., Carr, P., Lucey, S., Ramanan, D., et~al.: Argoverse: 3d tracking and
  forecasting with rich maps. In: Proceedings of the IEEE/CVF Conference on
  Computer Vision and Pattern Recognition. pp. 8748--8757 (2019)

\bibitem{chen2016monocular}
Chen, X., Kundu, K., Zhang, Z., Ma, H., Fidler, S., Urtasun, R.: Monocular 3d
  object detection for autonomous driving. In: Proceedings of the IEEE
  conference on computer vision and pattern recognition. pp. 2147--2156 (2016)

\bibitem{chen2019progressive}
Chen, Z., Zhang, J., Tao, D.: Progressive lidar adaptation for road detection.
  IEEE/CAA Journal of Automatica Sinica  \textbf{6}(3),  693--702 (2019)

\bibitem{chen2020puppeteergan}
Chen, Z., Wang, C., Yuan, B., Tao, D.: Puppeteergan: Arbitrary portrait
  animation with semantic-aware appearance transformation. In: Proceedings of
  the IEEE/CVF Conference on Computer Vision and Pattern Recognition. pp.
  13518--13527 (2020)

\bibitem{Chi_2021_CVPR}
Chi, C., Wang, Q., Hao, T., Guo, P., Yang, X.: Feature-level collaboration:
  Joint unsupervised learning of optical flow, stereo depth and camera motion.
  In: Proceedings of the IEEE/CVF Conference on Computer Vision and Pattern
  Recognition (CVPR). pp. 2463--2473 (June 2021)

\bibitem{clevert2015fast}
Clevert, D.A., Unterthiner, T., Hochreiter, S.: Fast and accurate deep network
  learning by exponential linear units (elus). arXiv preprint arXiv:1511.07289
  (2015)

\bibitem{dwivedi2021bird}
Dwivedi, I., Malla, S., Chen, Y.T., Dariush, B.: Bird’s eye view segmentation
  using lifted 2d semantic features. In: British Machine Vision Conference
  (BMVC). pp. 6985--6994 (2021)

\bibitem{fu2018deep}
Fu, H., Gong, M., Wang, C., Batmanghelich, K., Tao, D.: Deep ordinal regression
  network for monocular depth estimation. In: Proceedings of the IEEE
  Conference on Computer Vision and Pattern Recognition. pp. 2002--2011 (2018)

\bibitem{geiger2012we}
Geiger, A., Lenz, P., Urtasun, R.: Are we ready for autonomous driving? the
  kitti vision benchmark suite. In: Proceedings of the IEEE Conference on
  Computer Vision and Pattern Recognition. pp. 3354--3361. IEEE (2012)

\bibitem{godard2017unsupervised}
Godard, C., Mac~Aodha, O., Brostow, G.J.: Unsupervised monocular depth
  estimation with left-right consistency. In: Proceedings of the IEEE
  Conference on Computer Vision and Pattern Recognition. pp. 270--279 (2017)

\bibitem{monodepth2}
Godard, C., {Mac Aodha}, O., Firman, M., Brostow, G.J.: Digging into
  self-supervised monocular depth prediction  (October 2019)

\bibitem{hang2020human}
Hang, P., Lv, C., Xing, Y., Huang, C., Hu, Z.: Human-like decision making for
  autonomous driving: A noncooperative game theoretic approach. IEEE
  Transactions on Intelligent Transportation Systems  \textbf{22}(4),
  2076--2087 (2020)

\bibitem{he2016deep}
He, K., Zhang, X., Ren, S., Sun, J.: Deep residual learning for image
  recognition. In: Proceedings of the IEEE Conference on Computer Vision and
  Pattern Recognition. pp. 770--778 (2016)

\bibitem{fiery2021}
Hu, A., Murez, Z., Mohan, N., Dudas, S., Hawke, J., Badrinarayanan, V.,
  Cipolla, R., Kendall, A.: {FIERY}: Future instance segmentation in bird's-eye
  view from surround monocular cameras. In: Proceedings of the International
  Conference on Computer Vision ({ICCV}) (2021)

\bibitem{huang2021toward}
Huang, C., Lv, C., Hang, P., Xing, Y.: Toward safe and personalized autonomous
  driving: Decision-making and motion control with dpf and cdt techniques.
  IEEE/ASME Transactions on Mechatronics  \textbf{26}(2),  611--620 (2021)

\bibitem{kervadec2019boundary}
Kervadec, H., Bouchtiba, J., Desrosiers, C., Granger, E., Dolz, J., Ayed, I.B.:
  Boundary loss for highly unbalanced segmentation. In: International
  conference on medical imaging with deep learning. pp. 285--296. PMLR (2019)

\bibitem{kingma2014adam}
Kingma, D.P., Ba, J.: Adam: A method for stochastic optimization. arXiv
  preprint arXiv:1412.6980  (2014)

\bibitem{klingner2020self}
Klingner, M., Term{\"o}hlen, J.A., Mikolajczyk, J., Fingscheidt, T.:
  Self-supervised monocular depth estimation: Solving the dynamic object
  problem by semantic guidance. In: European Conference on Computer Vision. pp.
  582--600. Springer (2020)

\bibitem{li2018undeepvo}
Li, R., Wang, S., Long, Z., Gu, D.: Undeepvo: Monocular visual odometry through
  unsupervised deep learning. In: IEEE International Conference on Robotics and
  Automation. pp. 7286--7291. IEEE (2018)

\bibitem{lu2019monocular}
Lu, C., van~de Molengraft, M.J.G., Dubbelman, G.: Monocular semantic occupancy
  grid mapping with convolutional variational encoder--decoder networks. IEEE
  Robotics and Automation Letters  \textbf{4}(2),  445--452 (2019)

\bibitem{luo2021self}
Luo, C., Yang, X., Yuille, A.: Self-supervised pillar motion learning for
  autonomous driving. In: Proceedings of the IEEE/CVF Conference on Computer
  Vision and Pattern Recognition. pp. 3183--3192 (2021)

\bibitem{ma2020distance}
Ma, J., Wei, Z., Zhang, Y., Wang, Y., Lv, R., Zhu, C., Gaoxiang, C., Liu, J.,
  Peng, C., Wang, L., et~al.: How distance transform maps boost segmentation
  cnns: an empirical study. In: Medical Imaging with Deep Learning. pp.
  479--492. PMLR (2020)

\bibitem{mahjourian2018unsupervised}
Mahjourian, R., Wicke, M., Angelova, A.: Unsupervised learning of depth and
  ego-motion from monocular video using 3d geometric constraints. In:
  Proceedings of the IEEE Conference on Computer Vision and Pattern
  Recognition. pp. 5667--5675 (2018)

\bibitem{mallot1991inverse}
Mallot, H.A., B{\"u}lthoff, H.H., Little, J., Bohrer, S.: Inverse perspective
  mapping simplifies optical flow computation and obstacle detection.
  Biological cybernetics  \textbf{64}(3),  177--185 (1991)

\bibitem{mani2020monolayout}
Mani, K., Daga, S., Garg, S., Narasimhan, S.S., Krishna, M., Jatavallabhula,
  K.M.: Monolayout: Amodal scene layout from a single image. In: The IEEE
  Winter Conference on Applications of Computer Vision. pp. 1689--1697 (2020)

\bibitem{mccraith2020calibrating}
McCraith, R., Neumann, L., Vedaldi, A.: Calibrating self-supervised monocular
  depth estimation. arXiv preprint arXiv:2009.07714  (2020)

\bibitem{nister2004visual}
Nist{\'e}r, D., Naroditsky, O., Bergen, J.: Visual odometry. In: Proceedings of
  the 2004 IEEE Computer Society Conference on Computer Vision and Pattern
  Recognition, 2004. CVPR 2004. vol.~1, pp.~I--I. Ieee (2004)

\bibitem{pan2020cross}
Pan, B., Sun, J., Leung, H.Y.T., Andonian, A., Zhou, B.: Cross-view semantic
  segmentation for sensing surroundings. IEEE Robotics and Automation Letters
  \textbf{5}(3),  4867--4873 (2020)

\bibitem{paszke2017automatic}
Paszke, A., Gross, S., Chintala, S., Chanan, G., Yang, E., DeVito, Z., Lin, Z.,
  Desmaison, A., Antiga, L., Lerer, A.: Automatic differentiation in pytorch
  (2017)

\bibitem{philion2020lift}
Philion, J., Fidler, S.: Lift, splat, shoot: Encoding images from arbitrary
  camera rigs by implicitly unprojecting to 3d. In: European Conference on
  Computer Vision. pp. 194--210. Springer (2020)

\bibitem{Phillips_2021_CVPR}
Phillips, J., Martinez, J., Barsan, I.A., Casas, S., Sadat, A., Urtasun, R.:
  Deep multi-task learning for joint localization, perception, and prediction.
  In: Proceedings of the IEEE/CVF Conference on Computer Vision and Pattern
  Recognition (CVPR). pp. 4679--4689 (June 2021)

\bibitem{ranjan2019competitive}
Ranjan, A., Jampani, V., Balles, L., Kim, K., Sun, D., Wulff, J., Black, M.J.:
  Competitive collaboration: Joint unsupervised learning of depth, camera
  motion, optical flow and motion segmentation. In: Proceedings of the IEEE
  Conference on Computer Vision and Pattern Recognition. pp. 12240--12249
  (2019)

\bibitem{reading2021categorical}
Reading, C., Harakeh, A., Chae, J., Waslander, S.L.: Categorical depth
  distribution network for monocular 3d object detection. In: Proceedings of
  the IEEE/CVF Conference on Computer Vision and Pattern Recognition. pp.
  8555--8564 (2021)

\bibitem{roddick2020predicting}
Roddick, T., Cipolla, R.: Predicting semantic map representations from images
  using pyramid occupancy networks. In: Proceedings of the IEEE/CVF Conference
  on Computer Vision and Pattern Recognition. pp. 11138--11147 (2020)

\bibitem{roddick2018orthographic}
Roddick, T., Kendall, A., Cipolla, R.: Orthographic feature transform for
  monocular 3d object detection. arXiv preprint arXiv:1811.08188  (2018)

\bibitem{russakovsky2015imagenet}
Russakovsky, O., Deng, J., Su, H., Krause, J., Satheesh, S., Ma, S., Huang, Z.,
  Karpathy, A., Khosla, A., Bernstein, M., et~al.: Imagenet large scale visual
  recognition challenge. International journal of computer vision
  \textbf{115}(3),  211--252 (2015)

\bibitem{saha2021enabling}
Saha, A., Mendez, O., Russell, C., Bowden, R.: Enabling spatio-temporal
  aggregation in birds-eye-view vehicle estimation. In: 2021 IEEE International
  Conference on Robotics and Automation (ICRA). pp. 5133--5139. IEEE (2021)

\bibitem{schon2021mgnet}
Sch{\"o}n, M., Buchholz, M., Dietmayer, K.: Mgnet: Monocular geometric scene
  understanding for autonomous driving. In: Proceedings of the IEEE/CVF
  International Conference on Computer Vision. pp. 15804--15815 (2021)

\bibitem{shu2020featdepth}
Shu, C., Yu, K., Duan, Z., Yang, K.: Feature-metric loss for self-supervised
  learning of depth and egomotion. In: ECCV (2020)

\bibitem{simond2007obstacle}
Simond, N., Parent, M.: Obstacle detection from ipm and super-homography. In:
  2007 IEEE/RSJ International Conference on Intelligent Robots and Systems. pp.
  4283--4288. IEEE (2007)

\bibitem{Thavamani_2021_ICCV}
Thavamani, C., Li, M., Cebron, N., Ramanan, D.: Fovea: Foveated image
  magnification for autonomous navigation. In: Proceedings of the IEEE/CVF
  International Conference on Computer Vision (ICCV). pp. 15539--15548 (October
  2021)

\bibitem{torralba2002depth}
Torralba, A., Oliva, A.: Depth estimation from image structure. IEEE
  Transactions on pattern analysis and machine intelligence  \textbf{24}(9),
  1226--1238 (2002)

\bibitem{wagstaff2020self}
Wagstaff, B., Kelly, J.: Self-supervised scale recovery for monocular depth and
  egomotion estimation. arXiv preprint arXiv:2009.03787  (2020)

\bibitem{wang2018learning}
Wang, C., Miguel~Buenaposada, J., Zhu, R., Lucey, S.: Learning depth from
  monocular videos using direct methods. In: Proceedings of the IEEE Conference
  on Computer Vision and Pattern Recognition. pp. 2022--2030 (2018)

\bibitem{Wang_2021_CVPR}
Wang, H., Cai, P., Fan, R., Sun, Y., Liu, M.: End-to-end interactive prediction
  and planning with optical flow distillation for autonomous driving. In:
  Proceedings of the IEEE/CVF Conference on Computer Vision and Pattern
  Recognition (CVPR) Workshops. pp. 2229--2238 (June 2021)

\bibitem{wang2017deepvo}
Wang, S., Clark, R., Wen, H., Trigoni, N.: Deepvo: Towards end-to-end visual
  odometry with deep recurrent convolutional neural networks. In: 2017 IEEE
  international conference on robotics and automation (ICRA). pp. 2043--2050.
  IEEE (2017)

\bibitem{wang2021multi}
Wang, Y., Mao, Q., Zhu, H., Zhang, Y., Ji, J., Zhang, Y.: Multi-modal 3d object
  detection in autonomous driving: a survey. arXiv preprint arXiv:2106.12735
  (2021)

\bibitem{xu2021vitae}
Xu, Y., Zhang, Q., Zhang, J., Tao, D.: Vitae: Vision transformer advanced by
  exploring intrinsic inductive bias. Advances in Neural Information Processing
  Systems  \textbf{34},  28522--28535 (2021)

\bibitem{xue2020toward}
Xue, F., Zhuo, G., Huang, Z., Fu, W., Wu, Z., Ang, M.H.: Toward hierarchical
  self-supervised monocular absolute depth estimation for autonomous driving
  applications. In: 2020 IEEE/RSJ International Conference on Intelligent
  Robots and Systems (IROS). pp. 2330--2337. IEEE (2020)

\bibitem{xue2020shape}
Xue, Y., Tang, H., Qiao, Z., Gong, G., Yin, Y., Qian, Z., Huang, C., Fan, W.,
  Huang, X.: Shape-aware organ segmentation by predicting signed distance maps.
  In: Proceedings of the AAAI Conference on Artificial Intelligence. vol.~34,
  pp. 12565--12572 (2020)

\bibitem{yang2021projecting}
Yang, W., Li, Q., Liu, W., Yu, Y., Ma, Y., He, S., Pan, J.: Projecting your
  view attentively: Monocular road scene layout estimation via cross-view
  transformation. In: Proceedings of the IEEE/CVF Conference on Computer Vision
  and Pattern Recognition. pp. 15536--15545 (2021)

\bibitem{yin2018geonet}
Yin, Z., Shi, J.: Geonet: Unsupervised learning of dense depth, optical flow
  and camera pose. In: Proceedings of the IEEE Conference on Computer Vision
  and Pattern Recognition. pp. 1983--1992 (2018)

\bibitem{zhang2020empowering}
Zhang, J., Tao, D.: Empowering things with intelligence: a survey of the
  progress, challenges, and opportunities in artificial intelligence of things.
  IEEE Internet of Things Journal  \textbf{8}(10),  7789--7817 (2020)

\bibitem{zhang2022vitaev2}
Zhang, Q., Xu, Y., Zhang, J., Tao, D.: Vitaev2: Vision transformer advanced by
  exploring inductive bias for image recognition and beyond. arXiv preprint
  arXiv:2202.10108  (2022)

\bibitem{zhang2022towards}
Zhang, S., Zhang, J., Tao, D.: Towards scale consistent monocular visual
  odometry by learning from the virtual world. In: 2022 IEEE International
  Conference on Robotics and Automation (ICRA) (2022)

\bibitem{zhao2020collaborative}
Zhao, H., Bian, W., Yuan, B., Tao, D.: Collaborative learning of depth
  estimation, visual odometry and camera relocalization from monocular videos.
  In: IJCAI. pp. 488--494 (2020)

\bibitem{zhou2017unsupervised}
Zhou, T., Brown, M., Snavely, N., Lowe, D.G.: Unsupervised learning of depth
  and ego-motion from video. In: Proceedings of the IEEE Conference on Computer
  Vision and Pattern Recognition. pp. 1851--1858 (2017)

\bibitem{Zhuang_2021_ICCV}
Zhuang, Z., Li, R., Jia, K., Wang, Q., Li, Y., Tan, M.: Perception-aware
  multi-sensor fusion for 3d lidar semantic segmentation. In: Proceedings of
  the IEEE/CVF International Conference on Computer Vision (ICCV). pp.
  16280--16290 (October 2021)

\bibitem{zou2020learning}
Zou, Y., Ji, P., Tran, Q.H., Huang, J.B., Chandraker, M.: Learning monocular
  visual odometry via self-supervised long-term modeling. In: European
  Conference on Computer Vision. pp. 710--727. Springer (2020)

\bibitem{zou2018df}
Zou, Y., Luo, Z., Huang, J.B.: Df-net: Unsupervised joint learning of depth and
  flow using cross-task consistency. In: The European Conference on Computer
  Vision. pp. 36--53 (2018)

\end{thebibliography}
\end{document}